%% file: arxiv.tex
\documentclass[10pt,twocolumn,letterpaper]{article}

\usepackage[pagenumbers]{cvpr} % To force page numbers, e.g. for an arXiv version

\input{preamble}
\definecolor{cvprblue}{rgb}{0.21,0.49,0.74}
\usepackage[pagebackref,breaklinks,colorlinks,allcolors=blue]{hyperref}

\usepackage{multirow}
\usepackage[table]{xcolor}
\usepackage{tcolorbox}
\tcbuselibrary{listings} % 

\newtcblisting{rawprompt}[1][]{
  colback=gray!20,
  colframe=gray!80,
  arc=2pt,
  boxrule=0.5pt,
  title=\textbf{Prompt},
  listing only,           % Treat content as raw text
  listing options={
    basicstyle=\ttfamily\scriptsize, % Typewriter font
    breaklines=true,            % Auto-wrap long lines
    breakatwhitespace=true,     % Clean breaks
    columns=fullflexible        % Better character spacing
  },
  #1
}

\def\paperID{2292} % *** Enter the Paper ID here
\def\confName{CVPR}
\def\confYear{2026}

\title{Visual Reasoning Tracer: Object-Level Grounded Reasoning Benchmark}

\author{
Haobo Yuan$^{1}$,
Yueyi Sun$^{2}$,
Yanwei Li$^{4}$,
Tao Zhang$^{5}$,
Xueqing Deng$^{1}$, \\
Henghui Ding$^{6}$,
Lu Qi$^{5}$,
Anran Wang$^{3}$,
Xiangtai Li$^{3}$,
Ming-Hsuan Yang$^{1}$\\[2mm]
$^{1}$UC Merced \quad
$^{2}$PKU \quad
$^{3}$NTU \quad
$^{4}$CUHK \quad
$^{5}$WHU \quad
$^{6}$FDU \\
{\small  haoboyuan@ucmerced.edu, xiangtai94@gmail.com} \\
}

\begin{document}
% \maketitle
\input{figs/fig_teaser}
\input{sec/0_abstract} 
\input{sec/1_intro}
\input{sec/2_task}
\input{sec/3_benchmark}
\input{sec/4_method}
\input{sec/5_exp}
\input{sec/6_related_work}
\input{sec/7_con}

{
    \small
    \bibliographystyle{ieeenat_fullname}
    \bibliography{main}
}

\appendix
\input{sec/X_suppl}

\end{document}

%% file: figs/fig_teaser.tex
% \begin{figure*}[t]
%     \centering
%     \includegraphics[width=1.\linewidth]{ICLR/figs/fig_teaser_iclr.pdf}
%     \caption{\textbf{Visual Reasoning Tracer (VRT).} Given an image and a question, the MLLM generates a step-by-step reasoning process. Each step is grounded by its corresponding visual tracer, making the model's decision-making process transparent.}
%     \label{fig:teaser}
% \end{figure*}

\twocolumn[{%
   \renewcommand\twocolumn[1][]{#1}%
   \maketitle
   \begin{center}
      \centering
      \captionsetup{type=figure}
\vspace{-9mm}
\includegraphics[width=1.\linewidth]{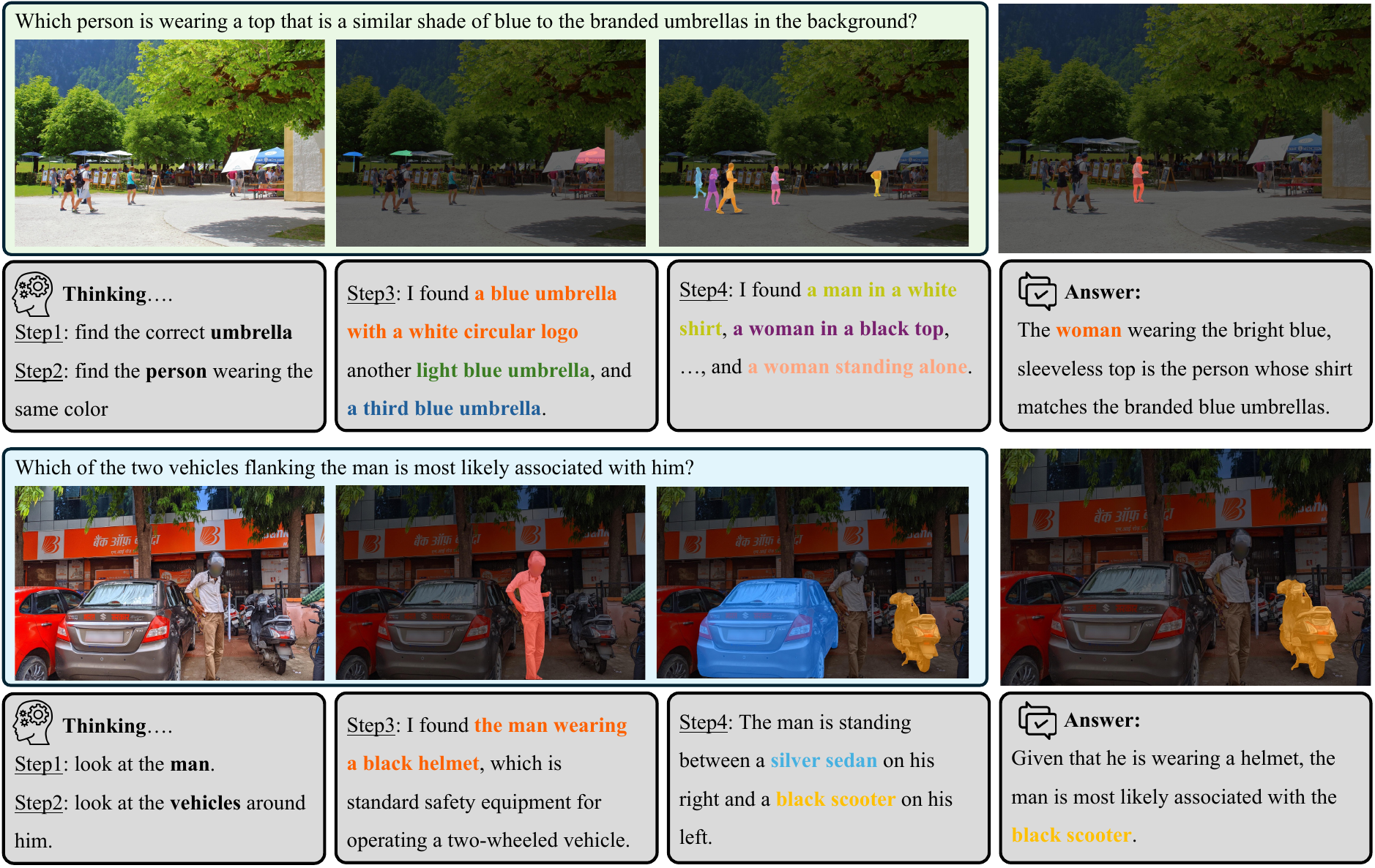}
      \captionof{figure}{\textbf{Visual Reasoning Tracer (VRT).} Given an image and a question, we ask an MLLM to generate a step-by-step reasoning process. Each step is grounded by its corresponding visual tracer, making the model’s decision-making process transparent and easy to understand. We use masks to ground each object in the scene. See more demos in the supplementary material.}
      \label{fig:teaser}
   \end{center}
}]

%% file: sec/0_abstract.tex
\begin{abstract}
Recent advances in Multimodal Large Language Models (MLLMs) have significantly improved performance on tasks such as visual grounding and visual question answering. However, the reasoning processes of these models remain largely opaque; they typically output only final predictions without revealing the intermediate steps or fine-grained evidence (e.g., pixels, locations) that lead to the result. This contrasts with human intelligence, which naturally operates through a chain of visual reasoning. To address this limitation, we introduce the Visual Reasoning Tracer (VRT) task, which requires models to not only localize the target object but also explicitly predict the intermediate objects that form the reasoning path. To advance research in this area, we contribute: (1) VRT-Bench, a human-annotated benchmark for evaluating visual reasoning; (2) a new metric for assessing the quality of reasoning traces; and (3) VRT-80k, a large-scale dataset for reasoning model training. Our experiments reveal that while existing models often produce the correct final output, they struggle to ground their intermediate reasoning. In contrast, models trained on VRT-80k achieve substantial improvements in tracing the reasoning path. All benchmarks, and code are available at \href{https://harboryuan.github.io/visual-reasoning-tracer}{here}.
\end{abstract}

%% file: sec/1_intro.tex
\section{Introduction}
\label{sec:intro}

Driven by large-scale pre-training and supervised fine-tuning, state-of-the-art (SOTA) multimodal large language models (MLLMs)~\citep{comanici2025gemini,qwen3vl,openai_o1} have demonstrated strong performance across diverse tasks, including Visual Question Answering (VQA)~\cite{liu2023llava,chen2024internvl}, Visual Grounding~\cite{bai2025qwen2,you2024ferret,zhang2024ferret,xiao2024florence,xiao2024oneref,zhang2024llava,liu2025seg,yuan2025sa2va}, and Referring Expression Segmentation (RES)~\cite{lai2024lisa,yuan2024osprey,zhang2024omg,liu2025seg,kang2025your}. While they have become adept at generating rich descriptions of visual content, this success raises a key challenge: can MLLMs move beyond simple description to precisely ground all mentioned elements, including both foreground objects and background areas, within an image, as humans do? This question probes whether these models truly understand the spatial and contextual relationships they describe.

Humans typically locate objects in complex scenes by sequentially identifying salient visual cues that establish a reasoning context. In a similar vein, OpenAI introduced the o3 family of models~\cite{openai2025introducing}. The o3 model and subsequent works~\cite{su2025thinking,wang2025traceable,guo2025beyond,liu2025small} demonstrate the ability to ``think with images'' by actively manipulating visual inputs (e.g., cropping and zooming) as part of their internal reasoning process. However, a critical gap exists in current evaluation methods.
Most benchmarks~\cite{guo2025beyond,shao2024visual,shen2025zoomeye,yan2025mmcr} focus on the accuracy of the final prediction, such as producing a correct textual answer to a visual question.
Consequently, the intermediate reasoning processes, especially the step-by-step, fine-grained pixel grounding that leads to a result, remain largely opaque.
Although SOTA models~\cite{yuan2025sa2va,zhang2025pixel,qwen3vl,comanici2025gemini} often produce the correct final output, their ability to faithfully ground their intermediate reasoning remains largely untested, revealing a significant limitation in how MLLMs are currently assessed.

To fill the gap, we propose the \textbf{VRT} (\textbf{V}isual \textbf{R}easoning \textbf{T}racer) task. The VRT task requires models to explicitly localize and predict the intermediate objects that form a reasoning path, not just the final target.
Our VRT task focuses on the traceable reasoning process. Specifically, VRT asks an MLLM to output a visual trace, a sequence of spatially grounded segmentation masks aligned with each reasoning step in its textual explanation.
As illustrated in~\Cref{fig:teaser}, given an image–question pair, the model must progressively highlight the regions it relies on (e.g., each telephone booth) while verbalizing the corresponding thought process.
This design bridges the gap between textual reasoning and perceptual grounding, making the model’s decision pathway transparent and interpretable.

To facilitate the evaluation of VRT, we introduce \textbf{VRT-Bench}, a meticulously human-annotated benchmark designed to test a model's ability to produce faithful reasoning paths. However, existing metrics are insufficient, as they typically focus only on the accuracy of the final target prediction, ignoring the intermediate steps. To address this, we propose a novel metric specifically designed to assess the fidelity of the entire visual trace, measuring how accurately the sequence of grounded masks aligns with the model's articulated reasoning.
Finally, to equip models with this new capability, we construct VRT-80k, a large-scale training dataset. These resources collectively provide the first comprehensive framework for developing and measuring transparent, step-by-step visual grounding in MLLMs.

We evaluate several SOTA MLLMs on VRT-Bench. Our benchmark findings highlight a critical disconnect: while many advanced reasoning models can produce the correct final textual answer, they fail to generate the intermediate visual reasoning trace. This confirms that their internal grounding remains opaque, even when the final output is correct. Our experiments also demonstrate a clear path forward. We show that by applying supervised fine-tuning (SFT) and Reinforcement Learning (RL) on our large-scale VRT-80k dataset, an MLLM with inherent pixel-wise grounding capabilities can successfully learn to produce these explicit, step-by-step visual traces, achieving strong performance on our new task. The main contributions of this work are as follows:
\begin{itemize}[leftmargin=20pt, topsep=0pt, itemsep=1pt, partopsep=1pt, parsep=1pt]
\item We introduce the Visual Reasoning Tracer, a novel task that requires models to jointly predict the reasoning process and corresponding context masks, thereby enhancing the interpretability of MLLM reasoning. To our knowledge, this is the first study to explore interactive, joint outputs of reasoning text and visual masks.
\item We annotate VRT-Bench to evaluate both reasoning and dense grounding capabilities. In addition, we propose a novel metric, Visual Quality (VQ), to jointly assess the quality of the reasoning process and final outputs. We construct \textbf{VRT-80k}, a large-scale, high-quality training dataset for developing and evaluating models on this task.
\item We conduct a comprehensive evaluation of SOTA MLLMs on VRT-Bench. Our analysis provides the first empirical evidence that current models, while often producing correct final answers, fail to ground their intermediate reasoning steps. We further demonstrate that with VRT-80k, models can successfully learn this capability and achieve strong performance, setting a new baseline for interpretable visual reasoning.
\end{itemize}

%% file: sec/2_task.tex
\section{Visual Reasoning Tracer}
\label{sec:task}

\subsection{Task Definition}\label{sec:task_def}
\noindent
\textbf{Motivation.} While recent MLLMs perform strongly on a wide range of tasks, they often lack transparent reasoning, making it difficult to assess whether their predictions are reasonable or truly reflect what is in the images. 
Although some previous works~\cite{shao2024visual,liu2025seg,qwen3vl} implement visual reasoning, the reasoning process is purely linguistic and is not sufficiently grounded in a visual trace.
In contrast to text-only reasoning, VRT enables a more interpretable and verifiable reasoning process. 
By requiring models to align each reasoning step with a visual segmentation mask, VRT bridges the gap between visual traces and textual explanations.

\noindent
\textbf{Task Formulation.} VRT requires a model to perform step-by-step reasoning over visual content by grounding intermediate reasoning steps. These reasoning steps are expected to be solidly grounded in a visual trace. 
Specifically, given an image $\mathbf{I}$ and a textual prompt $\mathbf{Q}$, where $\mathbf{Q}$ is a question that requires the model to use object(s) in the image to answer the question, the model is expected to generate a multi-modal reasoning trace:
\begin{equation}
    \mathbf{R} = \{(t_1, m_1), (t_2, m_2), ..., (t_n, m_n)\},
\end{equation}
where each $t_i$ is a natural language reasoning step that usually corresponds to a visual region in the image, 
and $m_i$ is a segmentation mask corresponding to the visual object(s) mentioned in $t_i$. 
The 2-tuple $v_i = (t_i, m_i)$ thus forms a multi-modal reasoning unit, linking linguistic inference with its corresponding visual trace in the image. 
The reasoning process may also include purely linguistic connections, which we ignore in this task formulation for simplicity. Given the multi-modal reasoning trace $\mathbf{R}$, the model outputs the answer $\mathbf{O} = \{A, m_{A}\}$. The model output contains both the language description $A$ and the corresponding object mask $m_A$. To represent an object accurately, we use a binary mask for each object; that is, $m_i \in \{0, 1\}^{H \times W}$ for the reasoning process and $m_{A} \in \{0, 1\}^{H \times W}$ for the final result. Using masks to represent objects allows the visual trace to be aligned at the pixel level.

\noindent 
\textbf{Task Scope and Capabilities.} To clearly define the scope, VRT concentrates on reasoning processes that are deeply rooted in visual trace, moving beyond abstract or external world knowledge. The core of this focus lies in understanding and articulating the relationships between objects within the image. To systematically evaluate this, we distinguish four primary types of reasoning capability that models must demonstrate: (1) Visual Details, which involves identifying specific attributes such as color, shape, or texture; (2) Location, which requires understanding the spatial arrangement and relative positioning of objects; (3) Function, which pertains to inferring the purpose or potential action of an object based on its visual context; and (4) Comparison, which involves assessing similarities or differences between two or more objects based on the preceding capabilities.

\input{figs/fig_bench_pipeline}
\subsection{Metrics}
\noindent

We propose new metrics inspired by \textbf{Panoptic Quality}~\cite{kirillov2019panoptic} for reasoning capability because existing evaluation methods for MLLM tasks are insufficient. Our design avoids reliance on problematic approaches like artificial formats and LLM-based evaluators.
Existing methods suffer from two major flaws. First, approaches like Multiple-Choice Questions (MCQ) rely on artificial formats that are not aligned with real-world user input, which does not provide predefined choices. Second, using another LLM to score the answer introduces a dependency on the evaluator model's own potential flaws.
Instead, our approach, centered on the VRT task, provides an objective solution. By requiring the model to produce grounded objects (i.e., segmentation masks) for all visual entities mentioned in its reasoning process, we can directly evaluate the fidelity of the reasoning trace against visual facts. This enables a fine-grained assessment of reasoning capability without resorting to artificial formats or a separate LLM evaluator.

\noindent
\textbf{Trace Matching.} To identify the correspondence between the mentioned objects in the prediction and the mentioned objects in the ground truth, we can naturally leverage the segmentation masks.
We establish this correspondence using bipartite matching, where the optimal pairing between predicted and ground-truth objects is determined by maximizing the total Intersection over Union (IoU) calculated from their segmentation masks.

Formally, let $M_{pred} = \{m_1, \dots, m_n\}$ be the set of all masks generated by the model (from the reasoning trace $\mathbf{R}$). Let $M_{gt} = \{m^{gt}_1, \dots, m^{gt}_k\}$ be the corresponding set of $k$ ground-truth masks. We construct a weight matrix $\mathbf{W}$ where each element $W_{ij}$ is the Intersection over Union (IoU) score between a predicted mask $m_i \in M_{pred}$ and a ground-truth mask $m^{gt}_j \in M_{gt}$:
$$
W_{ij} = \text{IoU}(m_i, m^{gt}_j) = \frac{|m_i \cap m^{gt}_j|}{|m_i \cup m^{gt}_j|}
$$
where $m_i, m^{gt}_j \in \{0, 1\}^{H \times W}$ and $|\cdot|$ denotes the count of non-zero pixels. We then solve this as a maximum weight bipartite matching~\cite{carion2020end} problem (e.g., using the Hungarian algorithm~\cite{kuhn1955hungarian}) to find the optimal pairing. A predicted object $m_i$ is considered successfully identified if it is matched to a ground-truth object $m^{gt}_j$ and their IoU score $W_{ij}$ exceeds a predefined threshold $\tau$ (e.g., $\tau = 0.5$).

\noindent
\textbf{Logic Quality.}
To evaluate the reasoning capacity, we divide the reasoning capacity into two levels. The first level is \textit{Logical Quality} (LQ). The LQ focuses on whether the model pays attention to the visual parts that are relevant to the question and expresses them in a recognizable description during the reasoning. We calculate the LQ as: 
\begin{equation}
LQ = \frac{|V_{\text{pred}} \cap V_{\text{GT}}|}{|V_{\text{GT}}|},
\end{equation}
where $ V_{\text{pred}} $ denotes the set of visual trace mentioned in the reasoning process, and $ V_{\text{GT}} $ denotes the ground-truth visual trace.

\noindent
\textbf{Visual Quality}
To evaluate the precision with which the model refers to visual content, we introduce a second level called \textit{Visual Quality} (VQ). While LQ measures whether the reasoning correctly mentions the necessary visual trace, VQ focuses on how accurately this visual trace is grounded spatially. Specifically, for cases where the visual trace is deemed correct (i.e., the visual trace exists in both prediction and ground truth), we compute the average Intersection over Union (IoU) between the predicted binary mask and the ground-truth binary mask. The VQ is calculated as:
\begin{equation}
VQ = \frac{1}{N} \sum_{i=1}^{N} \text{IoU}(m_{i}, m_{i}^{gt}),
\end{equation}
where $m_{i}$ and $m_{i}^{gt}$ denote the predicted and ground-truth masks for the i-th correctly identified visual trace. N is the total number of such clues across all samples with correctly identified visual trace. Together, LQ and VQ offer a comprehensive evaluation of the model's visual reasoning ability. LQ ensures that the reasoning process refers to the correct visual concepts. At the same time, VQ assesses the spatial precision of such references.

\input{figs/tex_fig_method}

%% file: figs/fig_bench_pipeline.tex
\begin{figure*}[t]
    \centering
    \includegraphics[width=1.\linewidth]{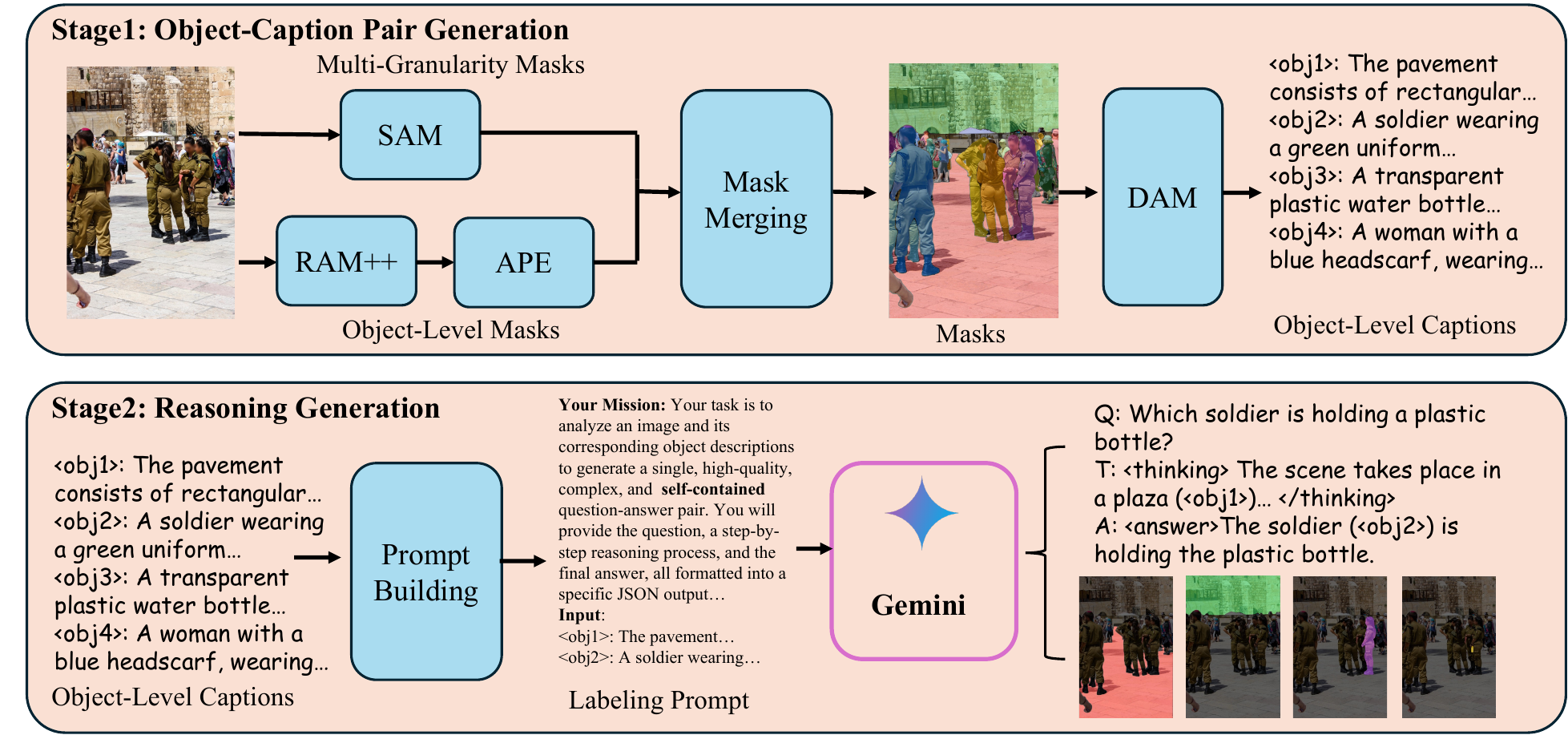}
    \caption{\textbf{The training data pipeline.} Our two-stage pipeline first generates object-caption pairs by segmenting an image with the Segment Anything Model (SAM) and RAM++/APE and describing each mask with the Describe Anything Model (DAM). In the second stage, these grounded captions are formatted into a prompt to guide Gemini in generating complex question-reasoning-answer data.}
    \label{fig:data_pip}
\end{figure*}

%% file: figs/tex_fig_method.tex
\begin{figure*}[t]
    \centering
    \includegraphics[width=1.\linewidth]{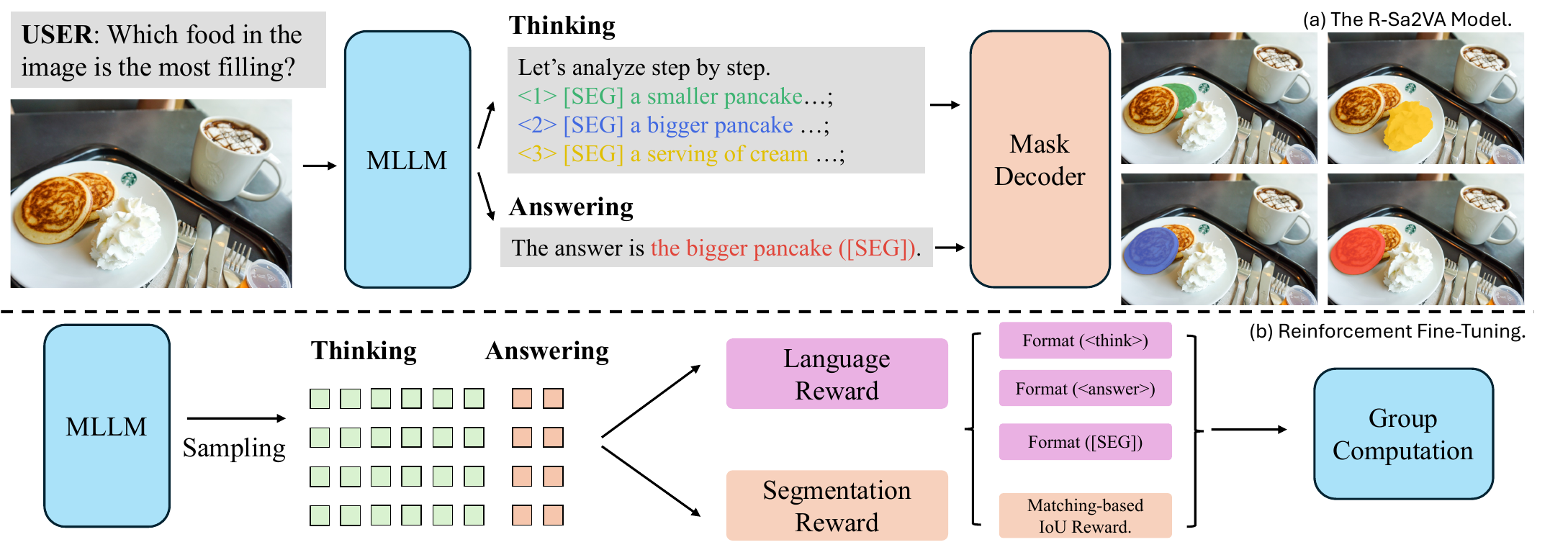}
    \caption{\textbf{Overview of the proposed R-Sa2VA framework.} (a) Inference Pipeline: Given a user query, the MLLM engages in an explicit visual reasoning process. It generates a ``Thinking'' sequence to identify and segment objects (e.g., distinguishing the "smaller pancake") before producing the final ``Answering'' sequence. A Mask Decoder translates the embedded [SEG] tokens into pixel-level masks, visually grounding the reasoning trace. (b) Reinforcement Fine-Tuning: To align the model's reasoning with accurate grounding, we employ a sampling-based optimization strategy. The model is trained using a dual-reward mechanism: a Language Reward that enforces structural consistency (e.g., correct usage of \texttt{\textless think\textgreater} tags) and a Matching-based IoU Reward that evaluates the quality of the generated masks.}
    \label{fig:method}
\end{figure*}

% ``\texttt{\textless think\textgreater}'' ``\texttt{\textless /think\textgreater}''

%% file: sec/3_benchmark.tex
\section{Data and Benchmark}
\label{sec:benchmark}

To train and evaluate models on the VRT task, we constructed two new datasets: \textbf{VRT-80k}, a large-scale, synthetically generated dataset for model training, and \textbf{VRT-Bench}, a high-quality, human-annotated benchmark for evaluation.

\subsection{VRT-80k: Large-Scale Training Set}
\label{sec:vrt-80k}

To generate a large volume of training data, we designed a two-stage data generation pipeline, as illustrated in~\Cref{fig:data_pip}. This pipeline first generates dense object-caption pairs and then uses a powerful MLLM to create complex reasoning question-answer pairs grounded in those objects.

\noindent
\textbf{Stage 1: Object-Caption Pair Generation.}
The first stage aims to densely segment an image and describe every resulting mask. Given an input image, we employ a multi-model approach for comprehensive object identification. We first use the \textbf{Segment Anything Model (SAM)}~\cite{kirillov2023segment} to produce multi-granularity masks, capturing everything from small parts to large background regions. Concurrently, we use object-level models like \textbf{RAM++}~\cite{zhang2024recognize} and \textbf{APE}~\cite{shen2024aligning} to generate precise masks for salient objects. These mask sets are then consolidated through a mask merging process to create a unified, non-redundant set of object masks. After that, each merged mask is fed into the \textbf{Describe Anything Model (DAM)}~\cite{lian2025describe} to generate a corresponding textual description. The output of this stage is a list of object-caption pairs (e.g., `(obj1, ``The pavement...'')`, `(obj2, ``A soldier...'')`), where each object identifier is linked to a segmentation mask and a description.

\noindent
\textbf{Stage 2: Reasoning Generation.}
In the second stage, we use the grounded object-caption pairs to generate complex reasoning data. We format the object captions from Stage 1 into a pure language prompt. As shown in~\Cref{fig:data_pip}, this prompt instructs an MLLM to perform a specific task: ``generate a simple, high-quality, complex, and self-contained question-answer pair'' using the provided objects (caption format) as input. This process yields a complete data sample: (Image, Question, Reasoning, Answer), where the reasoning and answer are directly grounded to the pixel-level masks generated in Stage 1. Following this pipeline, we generated \textbf{VRT-80k}, a large-scale dataset for training models on the VRT task.

\subsection{VRT-Bench: Annotated Evaluation Set}
\label{sec:vrt-bench}
While VRT-80k provides a large-scale training resource, synthetically generated data can contain noise or error samples. Thus, we need the human annotators to filter the correct data and group them into different reasoning categories. This categorization enables a fine-grained diagnosis of model limitations, as we discuss in~\Cref{sec:task_def}. We will provide more details in the supplementary material.

%% file: sec/4_method.tex
\input{tabs/tab_main}

\input{tabs/tab_stat}

\section{Method}
\label{sec:method}
\noindent
\textbf{Baseline.} We use the Sa2VA~\citep{yuan2025sa2va} as our baseline due to its simplicity and strong performance on referring segmentation tasks. 
Sa2VA is an MLLM that supports visual segmentation conditioned by natural language questions. 
Given an image and a textual prompt, Sa2VA can generate answers that contain a  ``[SEG]'' token which can be used to generate the mask referring to the object in question. 
It serves as the starting point for our training framework. However, similar to other existing MLLMs~\citep{lai2024lisa,zhang2024omg} that support segmentation, Sa2VA does not support complex concepts that may include multiple objects in the answer or visual reasoning trace. 
Therefore, we need to fine-tune Sa2VA to make it support the visual reasoning tracer.

\subsection{Our Method: R-Sa2VA}
R-Sa2VA is based on the Sa2VA~\citep{yuan2025sa2va} model. We load the checkpoint of Sa2VA and perform post-training to support VRT using the VRT-80k dataset.

\noindent
\textbf{Supervised Fine-Tuning}
In the training process, we directly conduct the Supervised Fine-Tuning (SFT) training following the training strategy of Sa2VA, which combines the following loss functions:
\begin{equation} 
    \mathcal{L}_{instruction} = \mathcal{L}_{text} + \mathcal{L}_{mask},\quad \mathcal{L}_{mask} = \mathcal{L}_{CE} + \mathcal{L}_{DICE}.\label{eq:sft} 
\end{equation}

\noindent
\textbf{Reinforcement Fine-Tuning.}
In the second training stage, we adopt an improved version of GRPO~\citep{shao2024deepseekmath,guo2025deepseek} for reinforcement fine-tuning. In reinforcement fine-tuning, the reward function design is essential since it guides the model's training direction. The reward function is as follows:
\begin{equation} 
    \mathcal{R} = \mathcal{R}_{format}^{t} + \mathcal{R}_{format}^{s} + \mathcal{R}_{iou}.
\end{equation}
The $\mathcal{R}_{format}^{t}$ is the thinking format reward. It guides the model to output the reasoning process with ``\texttt{\textless think\textgreater}'' ``\texttt{\textless /think\textgreater}'' enclosed. It also forces the model to output the answer in the ``\texttt{\textless answer\textgreater}'' ``\texttt{\textless /answer\textgreater}'' tags. If the output of the model does not meet the format, then there is no way to get the correct output. The $\mathcal{R}_{format}^{s}$ is the segmentation format reward. This reward is used to guide the model to output the ``[SEG]'' token. Only with the ``[SEG]'' tokens, the model can output valid masks. The first two rewards are similar to previous works~\citep{guo2025deepseek,liu2025seg}.

\noindent
\textbf{Matching-based IoU Reward.} The last part of the reward ($\mathcal{R}_{iou}$), we need to consider that the output may contain multiple instances. There may also be multiple objects in the ground truth. Unlike in the SFT process, where we have the determined order of the objects, the ground truth only contains objects included in the answer during the reinforcement fine-tuning. The answers with the swapped order of objects can all be deemed as correct. 
Thus, we propose to use a matching strategy that is invariant to the order of the objects. Specifically, we propose to use a bipartite matching algorithm to associate predicted masks with ground truth masks. For each predicted mask, we compute the Intersection over Union (IoU) with every ground truth mask, and then perform a one-to-one matching by selecting the pairs with the highest IoU scores iteratively, ensuring that each mask is only matched once. After the matching is completed, we compute the IoU for each matched pair and take the average as the final reward signal. 
Unmatched predicted masks and unmatched ground truth masks are considered as false positives and false negatives, respectively, and penalized accordingly. This method ensures that the reward is both order-invariant and sensitive to both segmentation accuracy and completeness of the prediction. The reward can be represented as follows:
\begin{equation} 
    \mathcal{R}_{iou} = \frac{1}{|\mathcal{M}|} \sum_{{p,g} \in \mathcal{M}}{\text{IoU}(p, g)} - \lambda\cdot|\mathcal{U}|,
\end{equation}
where $\mathcal{M}$ and $\mathcal{U}$ are matched and unmatched set for the bipartite matching. ${p,g}$ are matched objects from the predicted and ground truth sets. $\lambda$ is set to 0.1.

%% file: tabs/tab_main.tex
\begin{table*}[t]
\centering
\caption{\small{Quantitative evaluation of model performance across four core reasoning capabilities defined in the VRT benchmark. We report results on: (1) Comparison (\#comp), which assesses similarities or differences between objects; (2) Function (\#func), which infers the purpose or potential action of an object; (3) Location (\#loc), which tests the understanding of spatial arrangements; and (4) Visual Details (\#visf), which involves identifying specific attributes such as color, shape, or texture. For \textbf{R}easoning evaluation, we report LQ (\textbf{R-LQ}) and VQ (\textbf{R-VQ}). For \textbf{A}nswer evaluation, we report the mIoU (\textbf{A}).}}
\resizebox{1.\textwidth}{!}{
% Updated the column definition to 'ccc' for each metric group
\begin{tabular}{l|ccc|ccc|ccc|ccc|ccc}
\toprule[0.2em]
% --- Column Headers ---
% \multicolumn now spans 3 columns for each metric
\multirow{2}{*}{Model} & \multicolumn{3}{c|}{\#comp} & \multicolumn{3}{c|}{\#func} & \multicolumn{3}{c|}{\#loc} & \multicolumn{3}{c|}{\#visf} & \multicolumn{3}{c}{Overall} \\
% \cmidrules updated to span 3 columns each (e.g., 2-4, 5-7)
\cmidrule(lr){2-4} \cmidrule(lr){5-7} \cmidrule(lr){8-10} \cmidrule(lr){11-13} \cmidrule(lr){14-16}
% Second header row updated with the new sub-metrics
 & R-LQ & R-VQ & A & R-LQ & R-VQ & A & R-LQ & R-VQ & A & R-LQ & R-VQ & A & R-LQ & R-VQ & A \\
\midrule[0.1em]

% --- Data Rows ---
% \multicolumn for group headers now spans 16 columns (1 model + 15 data)
\multicolumn{16}{l}{\textit{(grounding models, box evaluation)}} \\
Gemini-2.5 Pro~\cite{comanici2025gemini} & 29.2 & 72.0 & 38.2 & 25.7 & 73.6 & 39.1 & 26.2 & 71.7 & 38.4 & 27.2 & 73.1 & 40.7 & 26.6 & 72.1 & 39.0 \\
Qwen3-VL-4B-Instruct~\cite{qwen3vl} & 6.9 & 85.5 & 53.5 & 9.3 & 88.2 & 51.2 & 12.5 & 85.4 & 52.5 & 12.4 & 85.0 & 54.2 & 10.2 & 86.2 & 50.6  \\
Qwen3-VL-8B-Instruct~\cite{qwen3vl} & 13.6 & 88.9 & 47.3 & 14.7 & 88.3 & 56.3 & 16.5 & 86.7 & 49.2 & 14.1 & 87.0 & 44.8 & 14.5 & 87.5 &  49.7 \\
\midrule

% Group 2: Segmentation Models
\multicolumn{16}{l}{\textit{(seg models, seg evaluation)}} \\
Gemini-2.5 Pro + SAM2 & 36.4 & 87.7 & 40.8 & 31.5 & 89.0 & 50.7 & 32.5 & 86.8 & 49.7 & 34.4 & 88.6 & 53.5 & 33.0 & 87.5 & 49.1 \\
Qwen3-VL-4B-Instruct + SAM2 & 6.6 & 95.5 & 57.3 & 9.3 & 92.0 &53.9 & 13.0 & 89.8 & 55.9 & 12.4 & 90.3 & 58.2 & 10.3 & 91.1 & 53.9 \\
Qwen3-VL-8B-Instruct + SAM2 & 13.6 & 93.8 & 48.9 & 15.2 & 91.2 & 58.5 & 16.3 & 91.2 & 51.2 & 14.3 & 91.7 & 45.9 & 14.7 & 91.7 & 51.7 \\
Sa2VA-Qwen3-4B~\cite{yuan2025sa2va} & 0.0 & 0.0 & 36.7 & 0.0 & 0.0 & 40.6 & 0.0 & 0.0 & 39.1 & 0.0 & 0.0 & 39.7 & 0.0 & 0.0 & 39.4 \\
\midrule

% Group 3: Reasoning Models
\multicolumn{16}{l}{\textit{(reasoning models, seg evaluation)}} \\
R-Sa2VA-InternVL-4B  &  60.5 & 87.9 & 51.6 & 62.6 & 88.1 & 54.6 & 60.2 & 86.8 & 49.6 & 60.6 & 87.8 & 56.5 & 61.2 & 87.5 & 54.0 \\
R-Sa2VA-Qwen3VL-4B & 66.6 & 87.5 & 61.7 & 68.9 & 87.9 & 60.9 & 65.5 & 86.2 & 55.3 & 64.0 & 87.6 & 63.0 & 66.3 & 87.3 & 59.5\\
% R-Sa2VA-Qwen3VL-4B-Thinking & 64.5 & 87.9 & 58.5 & 72.9 & 87.4 & 62.3 & 66.4 & 86.2 & 57.8 & 66.1 & 87.2 & 59.8 & 67.5 & 87.1 & 60.7\\
\bottomrule[0.1em]
\end{tabular}
}
\label{tab:main_R-VQesults}
\end{table*}

%% file: tabs/tab_stat.tex
\begin{table}[t]
\centering
\vspace{-2mm}
\caption{Count of samples for each category.}
\vspace{-3mm}
\label{tab:cnt_samples}
\resizebox{.9\linewidth}{!}{
\begin{tabular}{lcccccc}
\toprule[0.2em]
& Total & \#comp & \#func & \#loc & \#visf & multiple \\
\midrule
Count & 304 & 102 & 128 & 132 & 126 & 184 \\
\bottomrule[0.1em]
\end{tabular}
}
\vspace{-4mm}
\end{table}

%% file: sec/5_exp.tex
\section{Experiments}
\label{sec:exp}
\input{tabs/tab_abl_1_thinking_model}

\input{tabs/tab_abl_2_rl}

\input{tabs/tab_abl_3_refcoco}

\noindent
\textbf{Evaluation Datasets.}
We evaluate the performance on our proposed VRT-Bench. As detailed in~\Cref{tab:cnt_samples}, VRT-Bench comprises a total of 304 complex question-answer samples. These samples are grouped into four primary reasoning categories: comparison (\#comp), function (\#func), location (\#loc), and visual features (\#visf). The benchmark is well-distributed across these types, containing 102, 128, 132, and 126 samples for each category, respectively. Notably, the categories are not mutually exclusive; 184 samples are tagged with multiple reasoning types, underscoring the dataset's focus on complex, multi-step reasoning challenges.

\noindent
\textbf{Implementation Details.} 
We leverage the Sa2VA~\cite{yuan2025sa2va} framework to conduct continued training, utilizing the VRT-80k dataset for a duration of one epoch. The training process is distributed across 8 GPUs. In the SFT stage, we directly train 1 epoch on VRT-80k. In the RL stage, we select the hard samples from the VRT-80k and finetune 40 iters. See the supplementary material for more details.

% To align with the underlying architecture, we set the maximum sequence length to 8,192 tokens. The training process is distributed across 8 GPUs; with a per-device batch size of 2 and gradient accumulation steps set to 4, we achieve an effective global batch size of 64. Optimization is performed using AdamW with a learning rate of 4e-5. Additionally, we implement a linear warmup ratio of 0.05 and apply gradient clipping with a maximum norm of 1 to ensure training stability.

\noindent
\textbf{Comparison baselines.}
To provide a comprehensive evaluation, we select state-of-the-art MLLMs with reasoning and grounding capabilities, specifically Gemini-2.5 Pro~\cite{comanici2025gemini} and Qwen3-VL~\cite{qwen3vl}. For grounding methods, we directly use the box for evaluation, noting that converting VRT-Bench's segmentation masks to boxes serves only as an approximation. We also build baselines by combining these models with SAM-2 to evaluate segmentation results.

% While these models serve as strong baselines due to their robust reasoning and region-level grounding (bounding boxes), they face significant limitations: they are incapable of fine-grained, pixel-level grounding and cannot generate a visual trace of the reasoning process. In contrast, Sa2VA~\cite{yuan2025sa2va} distinguishes itself by supporting various pixel-level grounding tasks. However, similar to the baselines, Sa2VA also lacks the capability to generate a visual trace of the reasoning process.

\subsection{Main Results}
\label{sec:exp_main_results}

\noindent
\textbf{Benchmark Results.}
We evaluate the models across the four primary reasoning capabilities essential for the VRT task: Visual Details (\#visf), Location (\#loc), Function (\#func), and Comparison (\#comp). These categories rigorously test the model's ability to identify specific attributes (such as color or shape), understand spatial arrangements, infer object purposes, and assess similarities between objects, respectively. The results highlight a significant performance gap between model architectures. While standard grounding and segmentation baselines—such as Gemini-2.5 Pro and Qwen3-VL—maintain some general accuracy (A), they completely fail to generate valid reasoning outputs (scoring zero on R-LQ and R-VQ) across all four categories. In contrast, the R-Sa2VA-4B model demonstrates robust performance, successfully integrating visual perception with the deep reasoning required to articulate complex relationships and object attributes.

\subsection{Ablation Study and Analyze}\label{sec:exp_ablation}

\noindent\textbf{Language Thinking Model.}
In~\Cref{tab:abl_thinking_model}, we report our method with stronger language-wise thinking model. The results reveal that the R-Sa2VA-Qwen3VL-4B-Thinking model, which incorporates a stronger language thinking base model (Qwen3VL-4B-Thinking), demonstrates a notable improvement in specific reasoning capabilities. The most significant gain is observed in the function (\#func) category, where the Reasoning LQ (R-LQ) score increased by 4.0 points, from 68.9 to 72.9. However, this improvement is not uniformly salient across other categories. This suggests that while the "Thinking" model enhances abstract functional reasoning, its benefit is less pronounced for other tasks (e.g., visual details). Therefore, we hypothesize that introducing more samples of visual-traced reasoning data at a foundation-model scale could be a key strategy to further improve and generalize performance across all reasoning capabilities.

\noindent\textbf{Effect of Reinforcement Learning.}
In~\Cref{tab:abl_RL}, we report the effectiveness of Reinforcement Learning (RL). We observe that incorporating RL primarily enhances the model's logical reasoning capabilities, as evidenced by the consistent improvement in Reasoning Logic Quality (R-LQ) from 66.3 to 67.0. Interestingly, RL does not bring gains to the visual quality of the reasoning traces; the Reasoning Visual Quality (R-VQ) remains comparable or slightly lower than the SFT baseline, even with an additional segmentation loss. However, despite the lack of improvement in intermediate visual grounding, the final Answer performance (A) increases significantly (from 59.5 to 62.1). This suggests that the performance boost is driven by superior logical chain-of-thought, which guides the model to the correct final answer more effectively even without finer-grained segmentation in the reasoning steps.

\noindent\textbf{Effect of Joint Training.} In~\Cref{tab:abl_refcoco}, we investigate the impact of jointly training on our proposed VRT-80k dataset alongside standard referring segmentation datasets (RefCOCO, RefCOCO+, and RefCOCOg). The results demonstrate that the joint training strategy (Q3-4B-Joint) yields consistent improvements across standard benchmarks compared to the SFT baseline. However, we observe a slight performance decline on the VRT benchmark metrics (R-LQ, R-VQ, and A). We attribute this trade-off to the domain gap between the underlying visual data, as standard benchmarks utilize COCO images~\cite{lin2014microsoft} while VRT is constructed from the SA-1B dataset~\cite{kirillov2023segment}. Despite this minor quantitative fluctuation in reasoning scores, qualitative observations (\Cref{fig:vrt_vis}) indicate that the model benefits significantly from this regime; even without explicit reasoning annotations on COCO or reinforcement learning (RL), the joint training enables the model to effectively generate the visual evidence reasoning (VRT) process.

\input{figs/tex_fig_refcoco}

%% file: tabs/tab_abl_1_thinking_model.tex
\begin{table*}[t]
\centering
\caption{\small{\textbf{Ablation study on the base model.} We report Logic Quality (\textbf{R-LQ}), Visual Quality (\textbf{R-VQ}), and Answer mIoU (\textbf{A}.)}}
\resizebox{1.\textwidth}{!}{
% Updated the column definition to 'ccc' for each metric group
\begin{tabular}{l|ccc|ccc|ccc|ccc|ccc}
\toprule[0.2em]
% --- Column Headers ---
% \multicolumn now spans 3 columns for each metric
\multirow{2}{*}{Model} & \multicolumn{3}{c|}{\#comp} & \multicolumn{3}{c|}{\#func} & \multicolumn{3}{c|}{\#loc} & \multicolumn{3}{c|}{\#visf} & \multicolumn{3}{c}{Overall} \\
% \cmidrules updated to span 3 columns each (e.g., 2-4, 5-7)
\cmidrule(lr){2-4} \cmidrule(lr){5-7} \cmidrule(lr){8-10} \cmidrule(lr){11-13} \cmidrule(lr){14-16}
% Second header row updated with the new sub-metrics
 & R-LQ & R-VQ & A & R-LQ & R-VQ & A & R-LQ & R-VQ & A & R-LQ & R-VQ & A & R-LQ & R-VQ & A \\
\midrule[0.1em]
R-Sa2VA-Qwen3VL-4B-SFT & 66.6 & 87.5 & 61.7 & 68.9 & 87.9 & 60.9 & 65.5 & 86.2 & 55.3 & 64.0 & 87.6 & 63.0 & 66.3 & 87.3 & 59.5\\
R-Sa2VA-Qwen3VL-4B-Thinking & 64.5 & 87.9 & 58.5 & 72.9 & 87.4 & 62.3 & 66.4 & 86.2 & 57.8 & 66.1 & 87.2 & 59.8 & 67.5 & 87.1 & 60.7\\
\bottomrule[0.1em]
\end{tabular}
}
\label{tab:abl_thinking_model}
\end{table*}

%% file: tabs/tab_abl_2_rl.tex
\begin{table*}[t]
\centering
\caption{\small{\textbf{Effect of Reinforcement Learning.} We report Logic Quality (\textbf{R-LQ}), Visual Quality (\textbf{R-VQ}), and Answer mIoU (\textbf{A}.)}}
\resizebox{1.\textwidth}{!}{
% Updated the column definition to 'ccc' for each metric group
\begin{tabular}{l|ccc|ccc|ccc|ccc|ccc}
\toprule[0.2em]
% --- Column Headers ---
% \multicolumn now spans 3 columns for each metric
\multirow{2}{*}{Model} & \multicolumn{3}{c|}{\#comp} & \multicolumn{3}{c|}{\#func} & \multicolumn{3}{c|}{\#loc} & \multicolumn{3}{c|}{\#visf} & \multicolumn{3}{c}{Overall} \\
% \cmidrules updated to span 3 columns each (e.g., 2-4, 5-7)
\cmidrule(lr){2-4} \cmidrule(lr){5-7} \cmidrule(lr){8-10} \cmidrule(lr){11-13} \cmidrule(lr){14-16}
% Second header row updated with the new sub-metrics
 & R-LQ & R-VQ & A & R-LQ & R-VQ & A & R-LQ & R-VQ & A & R-LQ & R-VQ & A & R-LQ & R-VQ & A \\
\midrule[0.1em]
R-Sa2VA-Qwen3VL-4B-SFT & 66.6 & 87.5 & 61.7 & 68.9 & 87.9 & 60.9 & 65.5 & 86.2 & 55.3 & 64.0 & 87.6 & 63.0 & 66.3 & 87.3 & 59.5\\
\rowcolor{gray!20} +RL & 67.2 & 86.6 & 63.8 & 69.4 & 87.6 & 62.6 & 65.3 & 85.8 & 57.1 & 65.4 & 86.9 & 63.4 & 67.0 & 86.7 & 62.1\\
+RL + Seg Loss & 67.2 & 86.9 & 63.5 & 70.4 & 87.5 & 61.6 & 66.2 & 86.1 & 54.9 & 66.8 & 86.8 & 63.4 & 68.1 & 86.8 & 61.1\\
\bottomrule[0.1em]
\end{tabular}
}
\label{tab:abl_RL}
\end{table*}

%% file: tabs/tab_abl_3_refcoco.tex
\begin{table}[t]
\centering
\caption{\small{\textbf{Effect of joint training.} We report performance on standard referring segmentation datasets (RefCOCO, RefCOCO+, RefCOCOg, cIoU) alongside our VRT metrics: Logic Quality (\textbf{R-LQ}), Visual Quality (\textbf{R-VQ}), and Answer mIoU (\textbf{A}). For simplicity, Q3-4B means R-Sa2VA-Qwen3VL-4B.}}
\resizebox{1.\linewidth}{!}{
\begin{tabular}{l|ccc|ccc}
\toprule[0.2em]
% --- Top Header ---
% Group 1: Standard Datasets
% Group 2: VRT Benchmark Metrics
\multirow{2}{*}{Model} & \multicolumn{3}{c|}{RES} & \multicolumn{3}{c}{VRT Benchmark} \\
\cmidrule(lr){2-4} \cmidrule(lr){5-7} 
% --- Sub Header ---
% Columns for the specific datasets vs the specific metrics
 & \small{RefCOCO} & \small{RefCOCO+} & \small{RefCOCOg} & R-LQ & R-VQ & A \\
\midrule[0.1em]
Q3-4B-SFT   & 79.1&74.5 &	77.4 & 66.3 & 87.3 & 59.5 \\
Q3-4B-Joint & 81.7&77.6&80.0 & 65.9& 86.7 & 58.7 \\
\bottomrule[0.1em]
\end{tabular}
}
\label{tab:abl_refcoco}
\end{table}

%% file: figs/tex_fig_refcoco.tex
\begin{figure}[t]
    \centering
    \includegraphics[width=\linewidth]{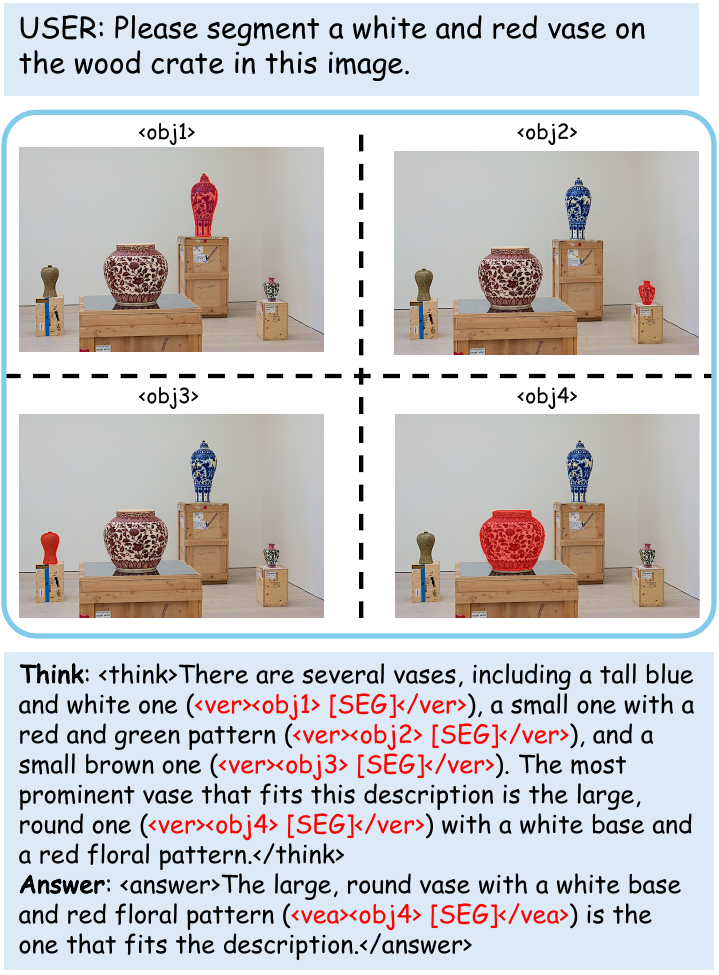}
    \caption{Visualization on RefCOCO dataset.}
    \label{fig:vrt_vis}
    \vspace{-3mm}
\end{figure}

%% file: sec/6_related_work.tex
\section{Related Work}
\label{sec:related_work}
% logic:
% 1. Large language model reasoning. 
% 2. Visual reasoning. 
% 3. Saptial-aware MLLMs.

\noindent
\textbf{Large Language Model Reasoning.} Recent advances in large language models (LLMs) have increasingly focused on enhancing reasoning capabilities, particularly through techniques such as Chain of Thought (CoT)~\citep{CoT_nips22}, which enable models to generate intermediate reasoning steps and improve performance on complex tasks. Building on this paradigm, models such as OpenAI-o1~\citep{openai_o1}, its improved variant o3, and DeepSeek-R1~\citep{guo2025deepseek} have attracted considerable attention for their advanced reasoning capabilities~\citep{xu2025towards}. LLM reasoning has been successfully applied to domains including mathematical problem solving~\citep{lightman2023let,yang2023leandojo,azerbayevllemma}, external tool use~\citep{schick2023toolformer}, robotics~\citep{kwon2024language}, complex task planning~\citep{wang2024planning}, and programming~\citep{hu2024can}, improving both performance and interoperability. In contrast to these prior works, our proposed VER task jointly investigates reasoning across both visual and textual modalities, extending beyond purely text-based reasoning.

\noindent
\textbf{Visual Reasoning.} Motivated by the success of LLM reasoning, recent studies~\citep{meng2025mm,yang2025r1,chen2025r1v,mmopenr1} have sought to extend reasoning capabilities to the visual domain. These works adapt CoT-style reasoning to multimodal LLMs (MLLMs), focusing primarily on image-level tasks~\citep{zhang2024cocot}, language-centric tasks~\citep{luan2024textcot,xie2025relationlmm,yao2023thinking}, and multimodal mathematical problem solving~\citep{sun2025mm,yang2025r1,chen2025r1v,wu2024mind}. Most approaches employ RL-like post-training on pre-trained MLLMs. More recently, some studies~\citep{kao2025think,liu2025seg} combine segmentation tasks with visual reasoning; however, these methods primarily focus on understanding the textual description prior to performing visual grounding. In contrast, our work unifies step-by-step visual grounding with text-based reasoning into a single, integrated process.

\noindent
\textbf{Spatial-Aware MLLMs.} Early MLLMs~\citep{liu2023llava} were limited to generating textual outputs. To enable richer interaction with the visual world—such as referring to objects while generating language—recent works~\citep{chen2023shikra,zhang2024llava,zhang2024ferret,wang2024qwen2,zhang2023gpt4roi,peng2023kosmos} introduced models capable of producing bounding boxes alongside text. To achieve more precise pixel-level references, subsequent studies~\citep{lai2024lisa,zhang2024omg,yan2024visa,xia2024gsva,yuan2025sa2va,zhang2025pixel} enabled MLLMs to generate segmentation masks. In this work, we go further by integrating visual trace with language reasoning, ensuring that every claim in the MLLM’s reasoning path is explicitly grounded in pixel-level observations. By emphasizing visual reasoning trace rather than merely providing visual answers, our approach produces a more transparent, interpretable, and reliable multimodal reasoning system.

%% file: sec/7_con.tex
\section{Conclusion}
\label{sec:conclusion}
This paper addresses a critical gap in current Multimodal Large Language Models (MLLMs): their opaque reasoning processes. While SOTA models can often produce correct final answers, they fail to expose or ground the intermediate steps that lead to these results. To address this, we introduce the Visual Reasoning Tracer (VRT) task, which requires models to generate an explicit, step-by-step reasoning path grounded by corresponding pixel-level segmentation masks. To support this new task, we contribute a suite of resources, including VRT-Bench, a human-annotated benchmark for evaluation; VRT-80k, a large-scale dataset for training; and novel metrics (LQ and VQ) to assess both the logical and visual grounding quality of the reasoning path. Our experiments demonstrate that while existing MLLMs are largely unable to produce such grounded traces, a model (R-Sa2VA) fine-tuned on our VRT-80k dataset can successfully acquire this capability and achieve strong performance.
By integrating visual evidence directly into language reasoning, this work provides a foundational framework for developing and evaluating MLLMs that are more interpretable and reliable, moving beyond merely producing correct answers to enabling verifiably grounded decision-making. Future directions include leveraging these grounded traces to enable model self-correction and extending the paradigm to dynamic video understanding.

%% file: sec/X_suppl.tex
\clearpage
\setcounter{page}{1}
\maketitlesupplementary

\renewcommand{\thetable}{A\arabic{table}}
\renewcommand{\thefigure}{A\arabic{figure}}
\setcounter{table}{0}
\setcounter{figure}{0}

\noindent
\textbf{Overview.}
This supplementary material provides additional details on the Visual Reasoning Tracer (VRT) beyond those in the main paper. We first elaborate on the construction of VRT-80k (\Cref{sec:VRT_data}), including the multi-stage data pipeline that combines object-level segmentation and captioning (\Cref{sec:multi_stage}), as well as the human labeling protocol and annotation tool (\Cref{sec:human_label}) used to curate VRT-Bench. We then present implementation and training details for R-Sa2VA (\Cref{sec:train_detail}), covering both supervised fine-tuning and reinforcement learning, together with the full set of hyperparameters. Next, we report extended experiments and ablations (\Cref{sec:more_exp}), analyzing the effect of example filtering, RL training dynamics, and providing rich qualitative visualizations of benchmark cases, model predictions, and filtered examples (\Cref{sec:more_vis}). Finally, we position VRT-Bench in relation to existing reasoning-aware segmentation and visual chain-of-thought benchmarks (\Cref{sec:diss}), and discuss future directions (\Cref{sec:future}). For better organization, we place all large figures at the end.

\section{More Details on our VRT datasets}\label{sec:VRT_data}
\subsection{More details on data pipeline}\label{sec:multi_stage}
\noindent\textbf{Stage 1: Object-Caption Pair Generation.} As illustrated in Figure 2 in the main paper, the foundation of our VRT-80k dataset is a set of high-quality, object-level mask-caption pairs. This paragraph details the multi-step process used to generate this data. Our pipeline integrates several visual foundation models (Recognize Anything (RAM++)~\cite{zhang2024recognize}, Aligning and Prompting Everything (APE)~\cite{shen2024aligning}, Segment Anything Model (SAM)~\cite{kirillov2023segment}, and Describe Anything Model (DAM)~\cite{lian2025describe}) and a post-processing workflow to ensure precision. To achieve comprehensive object coverage, we first generate two sets of masks. The first set is \textit{panoptic masks}. We use RAM++ to generate object tags for the image, which serve as an open-vocabulary list to guide APE in producing panoptic segmentation masks. At the same time, we generate \textit{multi-granularity masks} using the SAM model (everything mode). This step captures potential omissions from APE, such as background elements or objects not in APE's vocabulary. The union of these two mask sets provides excellent recall but contains many duplicates (e.g., a mask for ``person'' and another for ``person's upper body''). Subsequently, we apply a filtering process to merge these two mask types. During this process, two masks with an IoU higher than 0.5 are considered duplicates. We always keep the larger mask, as our focus is on object-level masks rather than part-level ones. Finally, we sample points from the filtered masks and use SAM again to generate the final mask. This process generally follows the methodology of DenseWorld-1M~\cite{li2025denseworld}, and we refer readers to their work for more details. However, we have one modification to the caption generation step, where we use the open-sourced DAM~\cite{lian2025describe}. 

\noindent\textbf{Stage 2: Reasoning Generation.} This stage is designed to address a key limitation in existing SOTA Multimodal Large Language Models (MLLMs). While MLLMs (such as Gemini 2.5 Pro~\cite{comanici2025gemini}) excelling at high-level, language-based reasoning , are not inherently trained to explicitly ground or ``point out'' the specific image regions corresponding to each step of their intermediate reasoning. Our objective is to ``link this gap'' by forcing the model to create a reasoning trace that is verifiably tied to visual evidence. To achieve this, we leverage the object-level captions generated in Stage 1 as the ``cookbook''. These object-caption pairs (e.g., (\texttt{\textless obj1\textgreater}, ``The pavement...'')) allow the MLLM to ``refer'' to a precise, pixel-level mask using only a simple language token. As shown in Figure 2 in the main paper, we inject these object-level captions into a ``Labeling Prompt'', which instructs the MLLM to generate a self-contained question-reasoning-answer triplet using the provided objects as input. This process yields a complete (Image, Question, Reasoning, Answer) sample, where the reasoning and answer are explicitly grounded to the pixel-level masks generated in Stage 1. We have provided the prompt of the reasoning generation in~\Cref{fig:prompt1,fig:prompt2,fig:prompt3}.

\subsection{More details on human labeling}\label{sec:human_label}
\noindent\textbf{Labeling Tool.}
Our labeling tool provides an integrated interface for creating and inspecting visual reasoning traces, as shown in~\Cref{fig:labeling}. The left panel displays the original input image, while the right panel shows the corresponding annotated segmentation masks with object IDs and colors for easy reference. Below, a text panel presents the generated question, step-by-step reasoning, and final answer, with inline object tags linking language to specific regions in the image. A key feature of our tool is that annotators can hover the mouse to selectively reveal specific masks in the image. This design allows annotators to efficiently verify both pixel-level masks and their alignment with the reasoning process, ensuring high-quality, object-grounded annotations.

\noindent\textbf{Labeling Guideline.}
For each candidate sample, annotators follow a four-stage protocol to ensure both human interpretability and precise pixel-level grounding. (1) The annotators verify whether the masks are accurate. If not, they directly drop the entire sample. (2) They then check whether the question is natural and understandable to a human viewer, and whether it can be answered purely from the given image and object masks. Questions that are ambiguous, poorly phrased, or not visually grounded (e.g., the answer is directly given in the mask list) are immediately rejected, and the entire sample is dropped. (3) Next, annotators carefully inspect the reasoning trace to confirm that it follows a coherent, step-by-step logic that a human expert would use, and that every object tag referenced in the text corresponds to the correct region in the annotated image. If the reasoning is incomplete, inconsistent, or relies on irrelevant or hallucinated objects, the annotator rewrites the reasoning in their own words and inserts only the minimal set of object tags that are strictly necessary to justify the answer. (4) Finally, they verify that the predicted answer is factually correct and fully supported by the validated reasoning and object tags; if not, the annotator provides the correct answer. We employ five expert annotators (all computer vision researchers with a PhD degree or studying as a PhD student), and only samples that pass all four stages are retained. After this multi-stage expert review, 304 out of the initial 1000 automatically generated samples are selected, yielding a benchmark with high accuracy, reliable reasoning traces, and strong alignment with human judgment.

\subsection{More Details on the Benchmark}
\noindent\textbf{Training / Evaluation Split.}
We note that the training and evaluation sets (before and after human labeling) are strictly separated. There are no overlapping images between the training and evaluation sets. In the training set, we do not apply a similar labeling process.

\begin{table}[t]
    \centering
    \small
    \begin{tabular}{l c}
        \toprule
        Hyperparameter & Value \\
        \midrule
        Max epochs & 1 \\
        Optimizer & AdamW \\
        Learning rate & $4\times 10^{-5}$ \\
        $\beta_1, \beta_2$ & 0.9, 0.999 \\
        Weight decay & 0.05 \\
        Max grad norm & 1.0 \\
        Warmup ratio & 0.05 \\
        Per-device batch size & 2 \\
        Gradient accumulation steps & 4 \\
        Number of GPUs & 8 (H100) \\
        Effective batch size & 64 \\
        Training time & $\sim$1.5 h \\
        \bottomrule
    \end{tabular}
    \caption{SFT hyperparameters for R-Sa2VA on VRT-80k.}
    \label{tab:rsa2va_sft_hparams}
\end{table}

\noindent\textbf{Filtered Examples.}
We present typical filtered cases from our VRT benchmark. In~\Cref{fig:fail1}, the Gemini generates a plausible answer, but the reasoning trace largely relies on superficial category cues (e.g., “holiday-themed”, “pumpkin toy”) rather than a meaningful multi-step logical analysis of the scene, indicating that the chain-of-thought can sometimes be unfaithful.
In~\Cref{fig:fail2}, the model correctly identifies the central skyscraper cluster described in the question. Still, the final answer mask is on two adjacent buildings, revealing limitations in fine-grained instance separation and precise object grounding in cluttered city scenes.
These issues arise from imperfections in the automatically collected VRT-80k datasets. 
Over half of the raw samples are filtered out in the VRT-Bench, suggesting that the VRT-80k is also likely noisy. Nevertheless, training on VRT-80k substantially boosts performance on our visual reasoning tracer benchmarks, suggesting that the dataset, while not perfect, already provides substantial supervision and that further improving the quality of the training datasets is a promising direction for future work.

% Main Experiments Supplementary.

\section{Implementation and Model Details}\label{sec:impl}
\subsection{Training Details.}\label{sec:train_detail}
Our R-Sa2VA training pipeline consists of two stages: a supervised fine-tuning (SFT) stage on VRT-80k, followed by a reinforcement learning (RL) stage on the filtered example.

\noindent\textbf{SFT.} 
We initialize R-Sa2VA from the public Sa2VA checkpoint and codebase~\cite{yuan2025sa2va} and apply LoRA-based parameter-efficient fine-tuning on the VRT-80k training split. We adopt a LoRA configuration with rank $r{=}128$, $\alpha{=}256$, and dropout of $0.05$. We train for a single epoch on 8 H100 GPUs with a per-device batch size of 2 and gradient accumulation steps of 4, leading to an effective batch size of 64. Training takes about 1.5 hours end-to-end. We use the AdamW optimizer with a learning rate of $4\times 10^{-5}$ (following the official Sa2VA setting), $(\beta_1,\beta_2)=(0.9,0.999)$, weight decay of 0.05, gradient clipping with a maximum norm of 1.0, and a warmup ratio of 0.05. The full set of SFT hyperparameters is summarized in \Cref{tab:rsa2va_sft_hparams}.

\begin{table}[t]
    \centering
    \small
    \begin{tabular}{l c}
        \toprule
        Hyperparameter & Value \\
        \midrule
        Max RL iterations & 40 \\
        Optimizer & AdamW \\
        Learning rate & $1\times 10^{-5}$ \\
        $\beta_1, \beta_2$ & 0.9, 0.999 \\
        Weight decay & 0.05 \\
        Max grad norm & 1.0 \\
        Warmup ratio & 0.05 \\
        Group size ($N$) & 4 \\
        Sampling temperature & 1.2 \\
        Top-$p$ & 0.9 \\
        Per-device batch size & 1 \\
        Gradient accumulation steps & 4 \\
        Number of GPUs & 8 (H100) \\
        Effective batch size & 32 \\
        Training time & $\sim$1 h \\
        \bottomrule
    \end{tabular}
    \caption{RL hyperparameters and configuration for R-Sa2VA.}
    \label{tab:rsa2va_rl_hparams}
\end{table}
\input{tabs/tab_abl4_filter}

\noindent\textbf{RL.} 
For the RL stage, we focus on the most ambiguous training examples identified via reward variance. Concretely, we first run inference with the SFT model on the entire VRT-80k training set, generating $N=4$ candidate answers per sample using the group-generation configuration described below. For each sample, we compute the task reward for all four candidates and take the standard deviation of these rewards as an uncertainty score. We then select the top 2k samples with the highest reward standard deviation as the RL subset. On this subset, we perform group-based RL fine-tuning with per-device batch size 1 and gradient accumulation steps of 4. We again use the AdamW optimizer and LORA, but reduce the learning rate to $1\times 10^{-5}$ and train for 40 RL iterations with the same $(\beta_1,\beta_2)$, weight decay, gradient clipping, and warmup settings as in SFT. During RL, each input is decoded with stochastic group generation: we sample four trajectories per input with a maximum of 1024 new tokens, temperature 1.2, top-$p=0.9$, and top-$k=0$. Hyperparameters are listed in~\Cref{tab:rsa2va_rl_hparams}.

\subsection{Model Details.}
Our implementation is built directly on the official Sa2VA model and training codebase~\cite{yuan2025sa2va}, which couples SAM2 with an MLLM for dense grounded understanding. We keep the original Sa2VA architecture unchanged, including the SAM2-based visual encoder, the [SEG] token interface, and the mask decoder, and only adapt the language head to emit our VRT-style outputs (step-by-step traces and final answers) while preserving Sa2VA’s input / output formats for segmentation. All data loading, optimization, and mask losses reuse the official training scripts and configurations. The Sa2VA repo exposes the base MLLM as a pluggable component, with multiple variants (e.g., InternVL2.5-/InternVL3-based 1B–26B models) provided in the model zoo. Switching the backbone is done by changing a single configuration entry without modifying any other part of the system. In our experiments, all R-Sa2VA variants (e.g., R-Sa2VA-InternVL2.5-4B, R-Sa2VA-Qwen3-VL-4B) share identical training pipelines and hyperparameters and differ only in this choice of base MLLM, ensuring that comparisons across backbones are fair and attributable to the underlying VLMs.

\section{More Experimental Results}\label{sec:more_exp}
In Sec.~5.2 of the main paper, we have already presented core ablation studies to validate the effectiveness of our design choices. In this section, we provide additional experimental results and extended ablations, offering a more comprehensive analysis.

\noindent\textbf{Ablation Study on the Example Filtering.} In~\Cref{sec:train_detail}, we introduce our example filtering strategy for RL. \Cref{tab:abl_filter} compares R-Sa2VA-Qwen3VL-4B-RL trained with and without this filtering. Removing the filter slightly degrades the overall answer mIoU (61.8 vs.\ 62.1) and leads to noticeable drops on the \#loc and \#visf subsets in both logic and visual quality, indicating that filtering is particularly useful for these more visual-related categories. The differences on \#comp and \#func are smaller, suggesting that filtering may not be good for these categories. For \#comp and \#func, filtering does not show a clear improvement, which we suspect is because they rely more on language logic and our reward is mainly focusing visual signals. Given these gains and the fact that filtering only affects the training data (with no additional training cost), we adopt the filtered setting as our default configuration for all other experiments.

\input{figs/fig_eval_results_comprehensive}
\noindent\textbf{Training Evaluation.}
\Cref{fig:eval_results_comp} analyzes how RL affects the VRT metrics over the course of training. Overall, the final answer scores (mIoU) initially improve compared to the SFT baseline, but can fluctuate and even slightly degrade if we continue RL for too long. In contrast, the reasoning and answer LQ curves show a clear and consistent improvement, indicating that RL effectively strengthens the logical structure of the reasoning traces. However, both reasoning and answer VQ decreases, and the answer mIoU also becomes less stable at later iterations, suggesting that the segmentation quality during reasoning is not reliably improved by our current reward design. Based on these observations, we adopt the model at 40 iterations as the default training setting (about 1 hour). Our hypothesis is that RL is primarily beneficial for reasoning, while it provides only weak or noisy signals for segmentation when no explicit mask supervision is used during RL; designing segmentation-aware rewards to better improve the masks is an interesting direction for future work.

\section{Visualization Results}\label{sec:more_vis}
In this section, we provide more visualization results, including our benchmark, model output, and visual comparison.

\subsection{VRT-Bench Visualizations}
\begin{figure*}
    \centering
    \includegraphics[width=1.\linewidth]{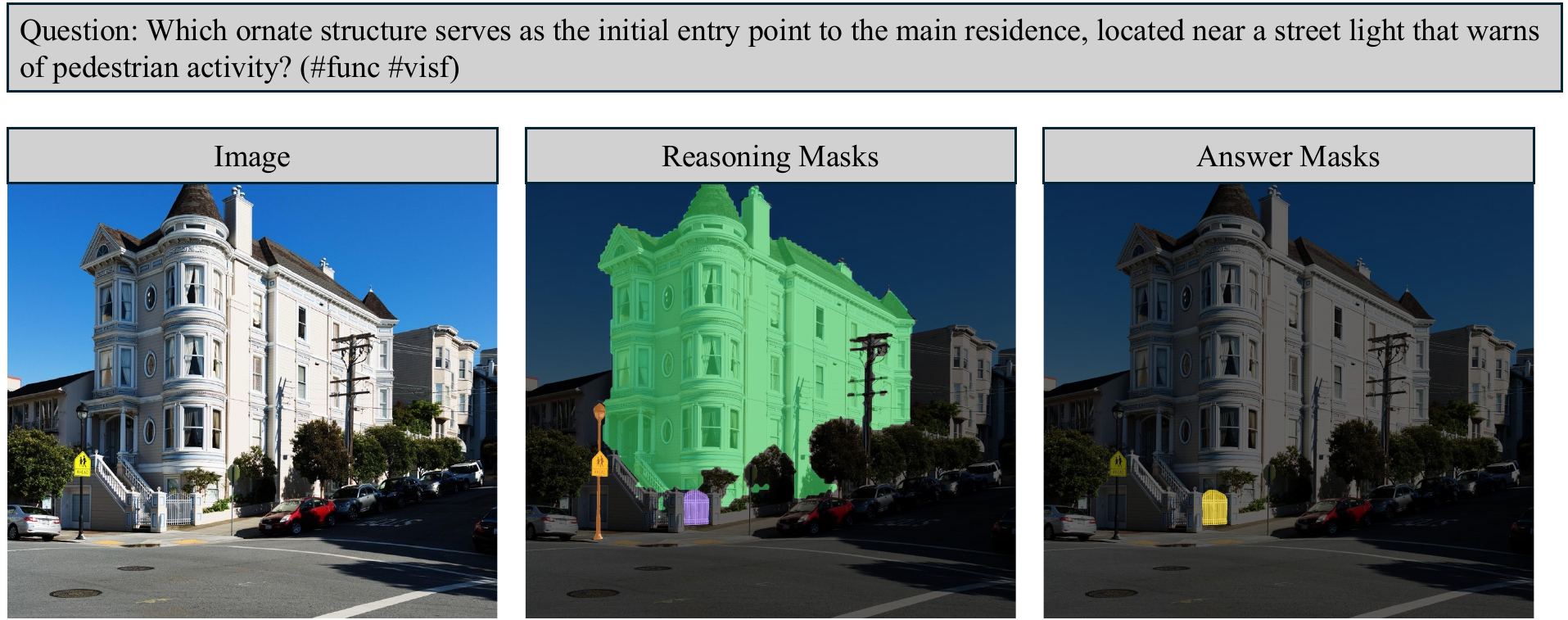}
    \caption{VRT-Bench Example 1.}
    \label{fig:vrtbench_1}
\end{figure*}

\begin{figure*}
    \centering
    \includegraphics[width=1.\linewidth]{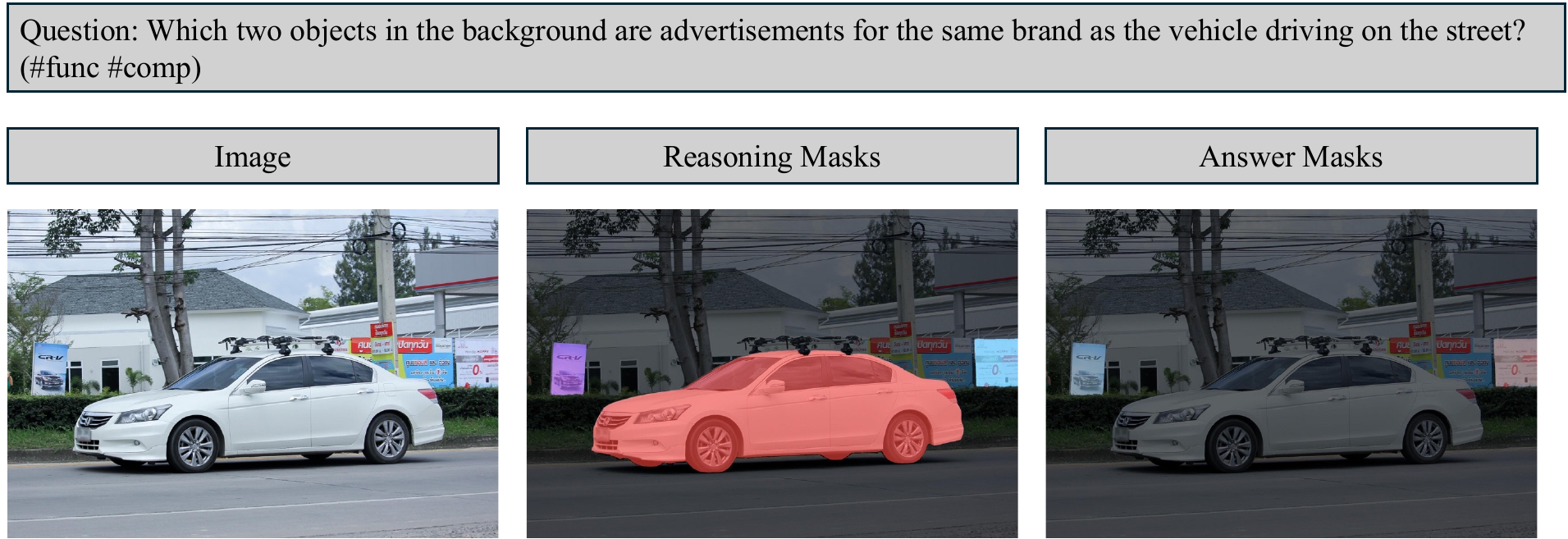}
    \caption{VRT-Bench Example 2.}
    \label{fig:vrtbench_2}
\end{figure*}

\begin{figure*}
    \centering
    \includegraphics[width=1.\linewidth]{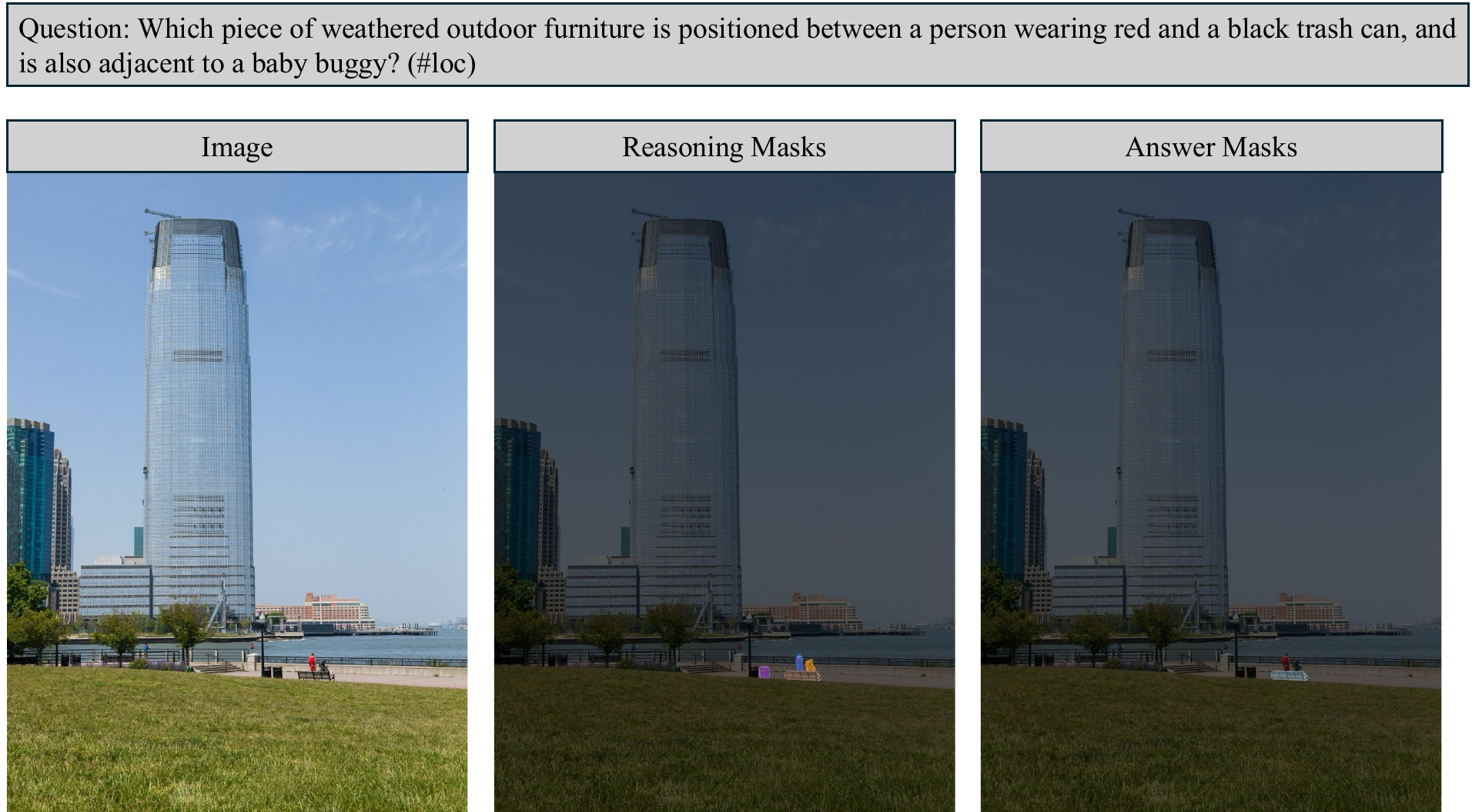}
    \caption{VRT-Bench Example 3.}
    \label{fig:vrtbench_3}
\end{figure*}

\begin{figure*}
    \centering
    \includegraphics[width=1.\linewidth]{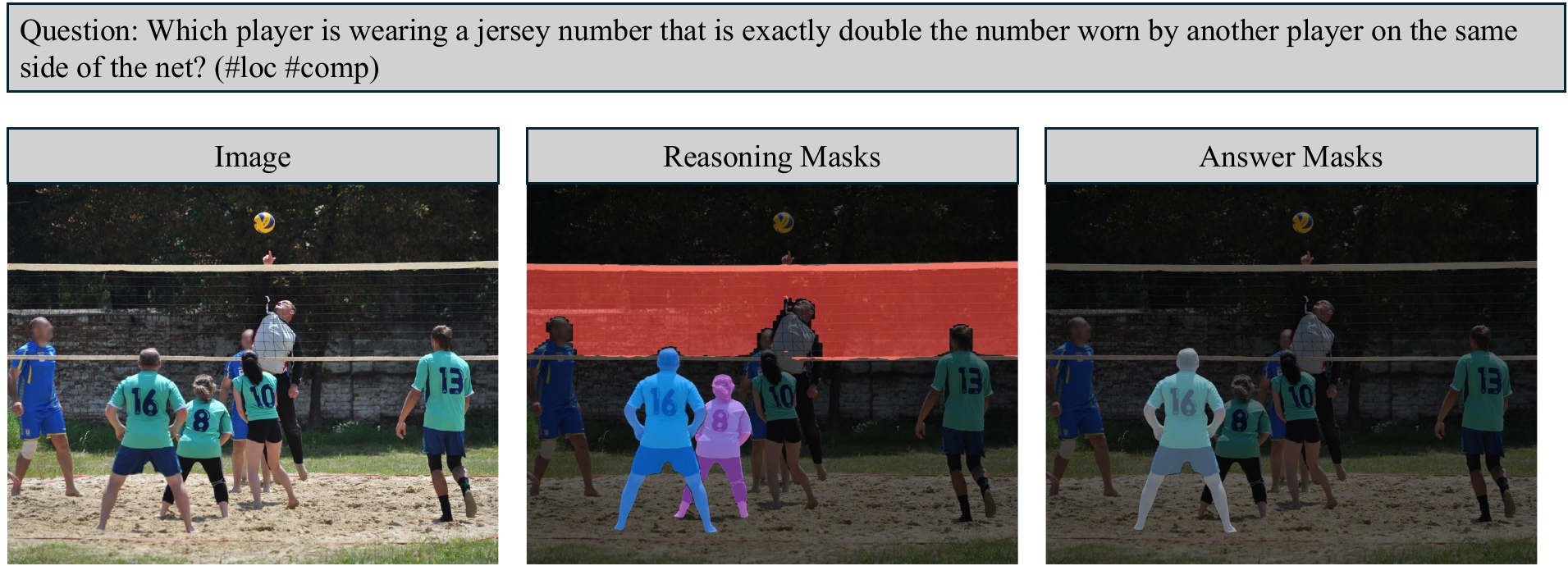}
    \caption{VRT-Bench Example 4.}
    \label{fig:vrtbench_4}
\end{figure*}

\Cref{fig:vrtbench_1,fig:vrtbench_2,fig:vrtbench_3,fig:vrtbench_4} illustrate four representative VRT-Bench test cases. \Cref{fig:vrtbench_1} targets functional (func) + visual-feature (visf) reasoning, asking the model to identify an ornate entrance structure near a pedestrian-warning street light, while~\Cref{fig:vrtbench_2} requires functional (func) + comparison reasoning (comp) to match a moving vehicle with background advertisements of the same brand. \Cref{fig:vrtbench_3} focuses on pure spatial reasoning, requiring the system to localize a specific piece of outdoor furniture based on its position relative to multiple surrounding objects. Finally, \Cref{fig:vrtbench_4} combines location and comparison reasoning, where the model must find the player whose jersey number is exactly double that of a teammate on the same side of the net. Together, these examples demonstrate that VRT-Bench includes diverse testing scenarios designed to systematically stress different reasoning dimensions.

\subsection{Visual Comparison}
\begin{figure*}[t]
    \centering
    \begin{minipage}{0.48\linewidth}
        \centering
    \includegraphics[width=1.\linewidth]{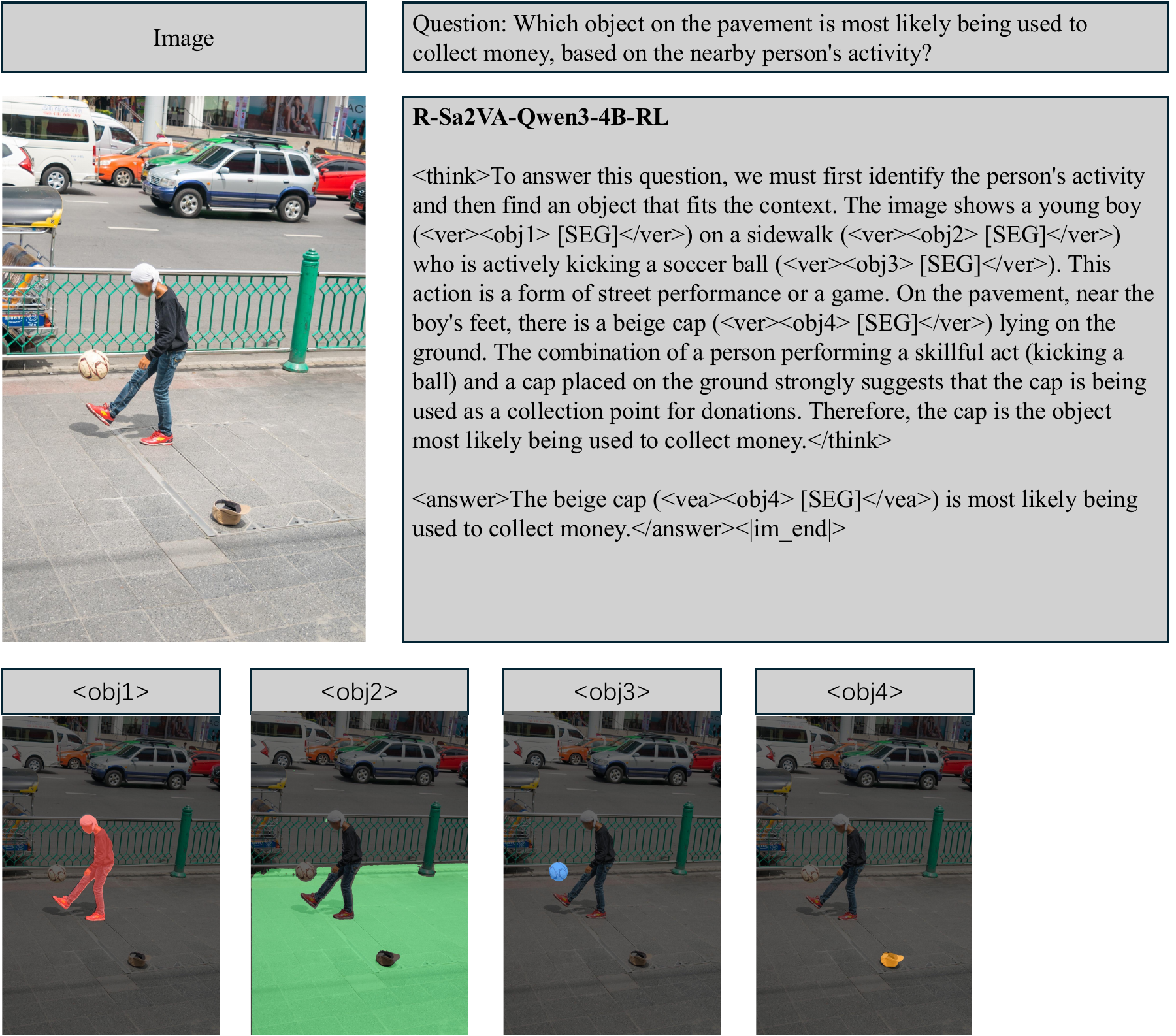}
    \caption{Visualization Results of our R-Sa2VA-Qwen3-4B-RL.}
    \label{fig:vis_sa1}
    \end{minipage}
    \hfill
    \begin{minipage}{0.48\linewidth}
        \centering
    \includegraphics[width=1.\linewidth]{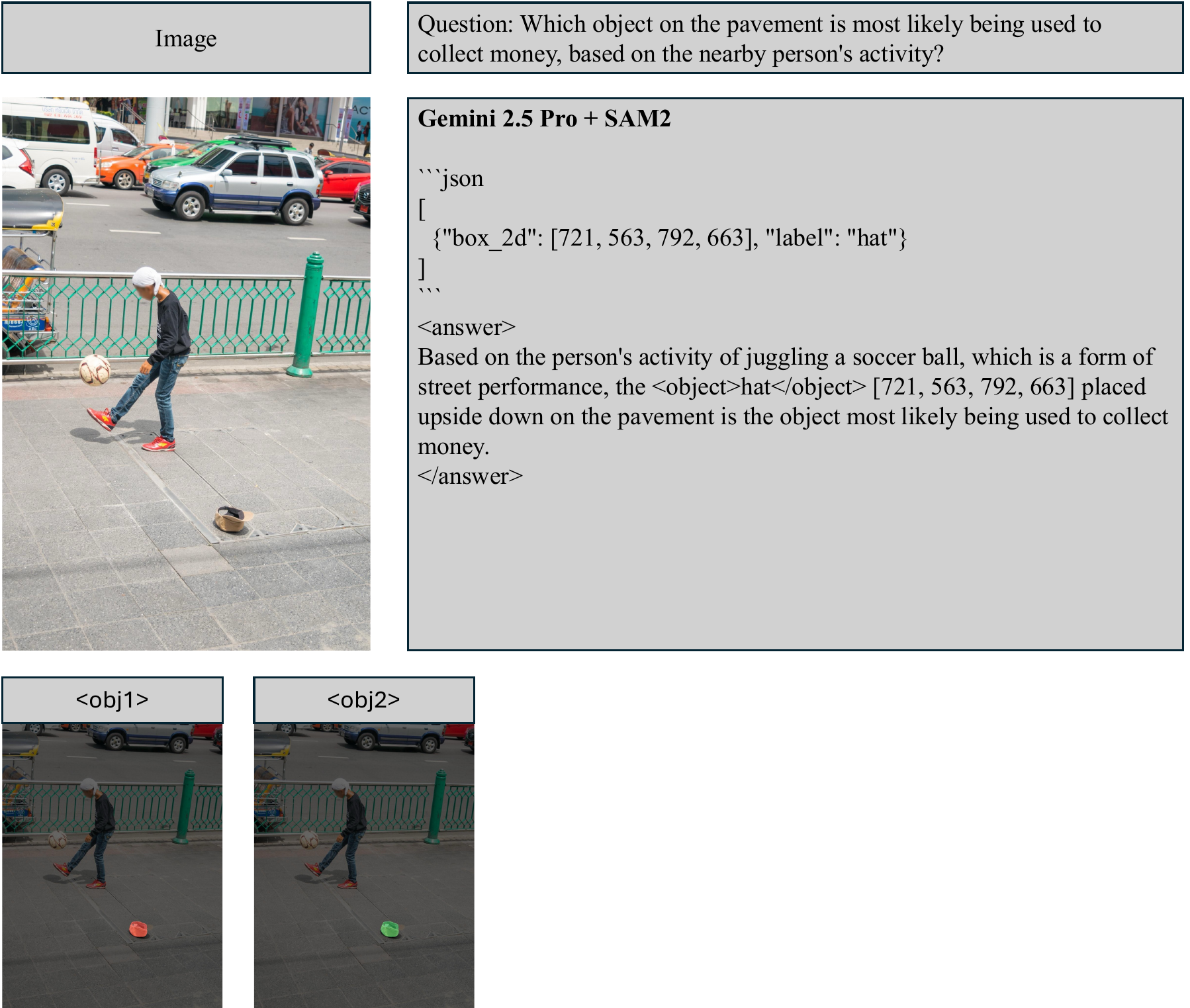}
    \caption{Visualization Results of Gemini 2.5 Pro.}
    \label{fig:vis_gp1}
    \end{minipage}
\end{figure*}

\begin{figure*}[t]
    \centering
    \begin{minipage}{0.48\linewidth}
        \centering
    \includegraphics[width=1.\linewidth]{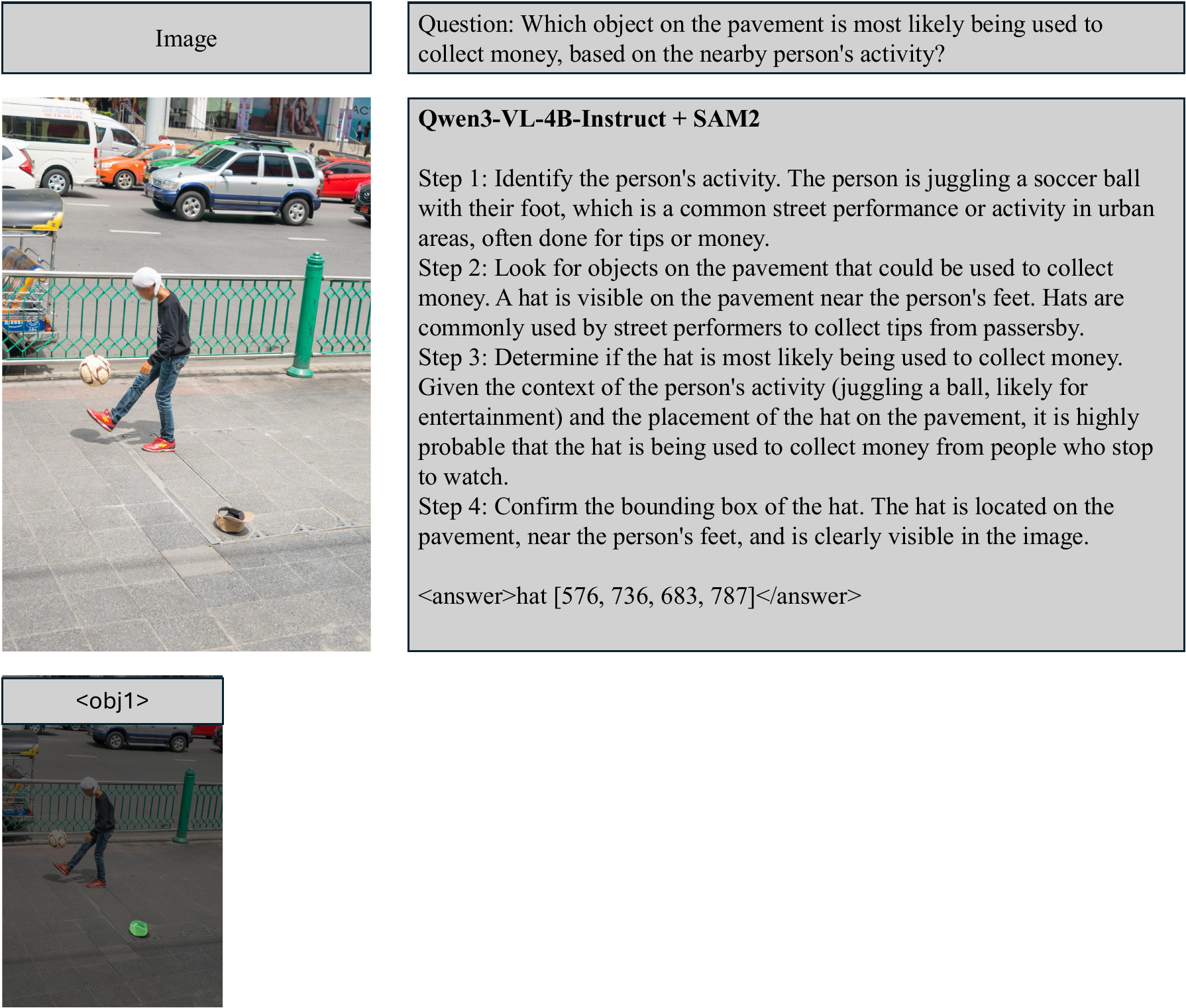}
    \caption{Visualization Results of Qwen3-4B.}
    \label{fig:vis_q1}
    \end{minipage}
    \hfill
    \begin{minipage}{0.48\linewidth}
        \centering
    \includegraphics[width=1.\linewidth]{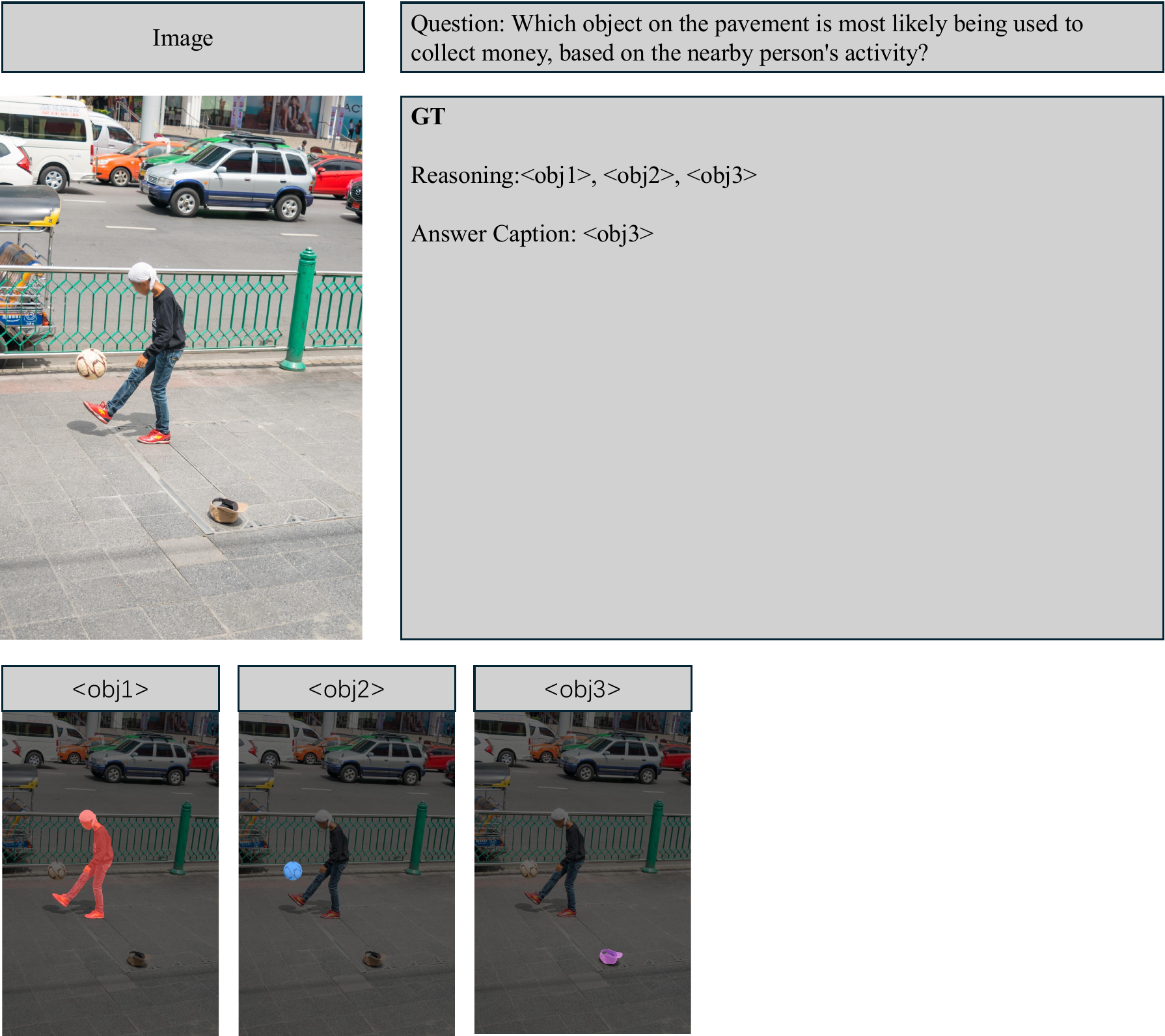}
    \caption{Visualization Results of ground truth.}
    \label{fig:vis_gt1}
    \end{minipage}
\end{figure*}

\begin{figure*}
        \centering
    \includegraphics[width=0.8\linewidth]{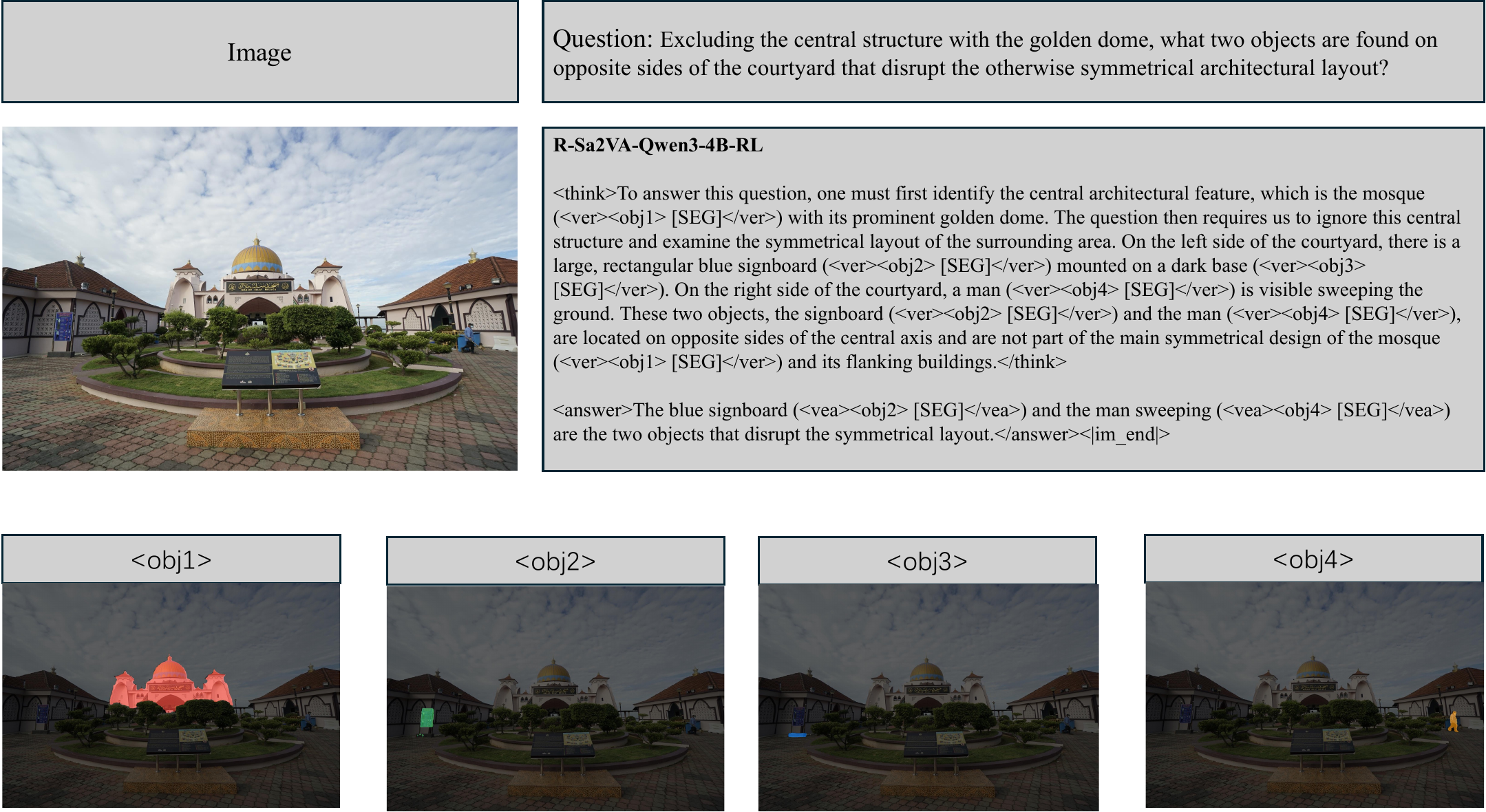}
    \caption{Visualization Results of our R-Sa2VA-Qwen3-4B-RL.}
    \label{fig:vis_sa2}
    \end{figure*}

\begin{figure*}
        \centering
    \includegraphics[width=0.8\linewidth]{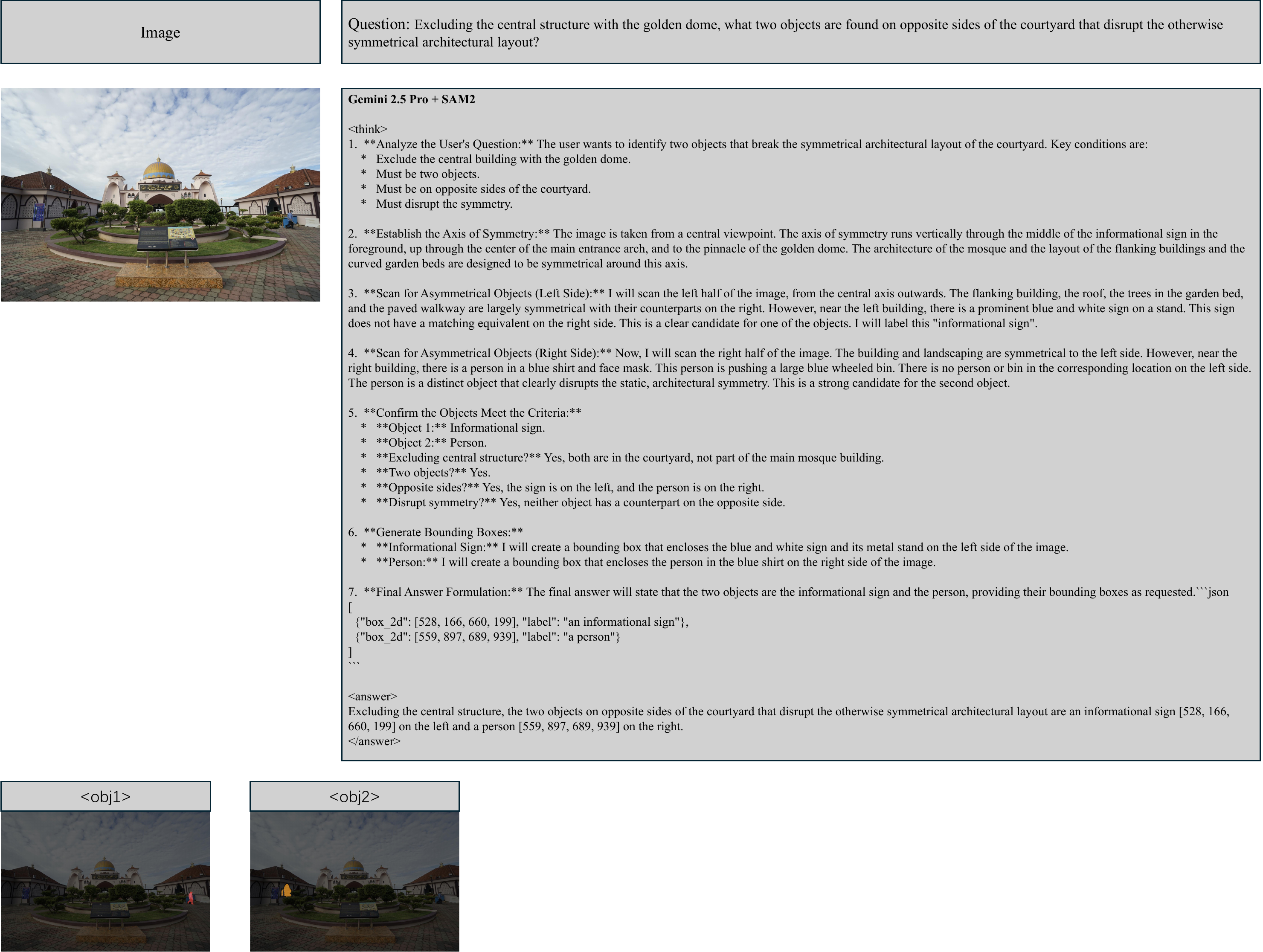}
    \caption{Visualization Results of Gemini 2.5 Pro + SAM2.}
    \label{fig:vis_gp2}
    \end{figure*}

\begin{figure*}
        \centering
    \includegraphics[width=.8\linewidth]{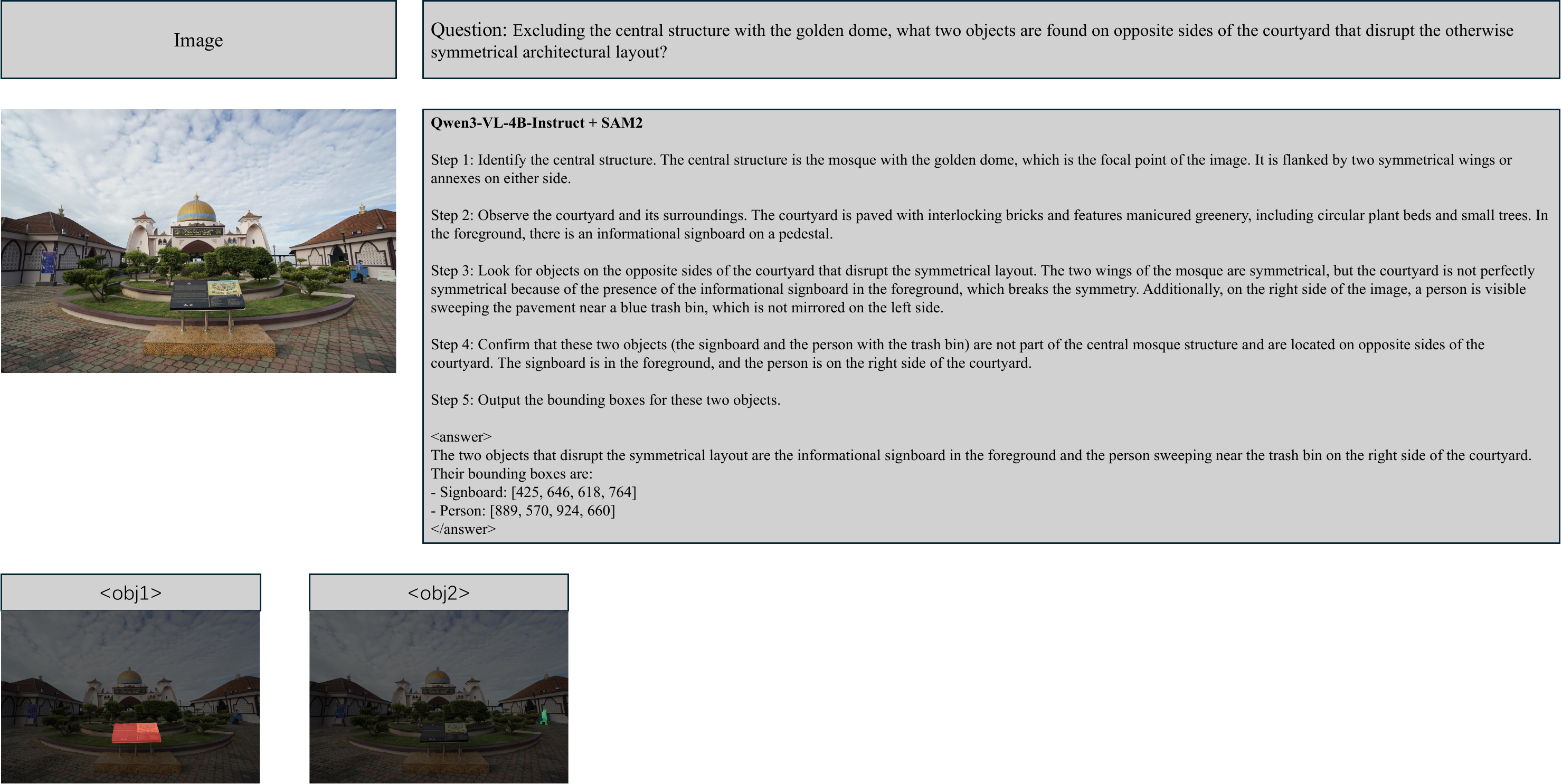}
    \caption{Visualization Results of Qwen3-VL-4B + SAM2.}
    \label{fig:vis_q2}
    \end{figure*}

\begin{figure*}
        \centering
    \includegraphics[width=.8\linewidth]{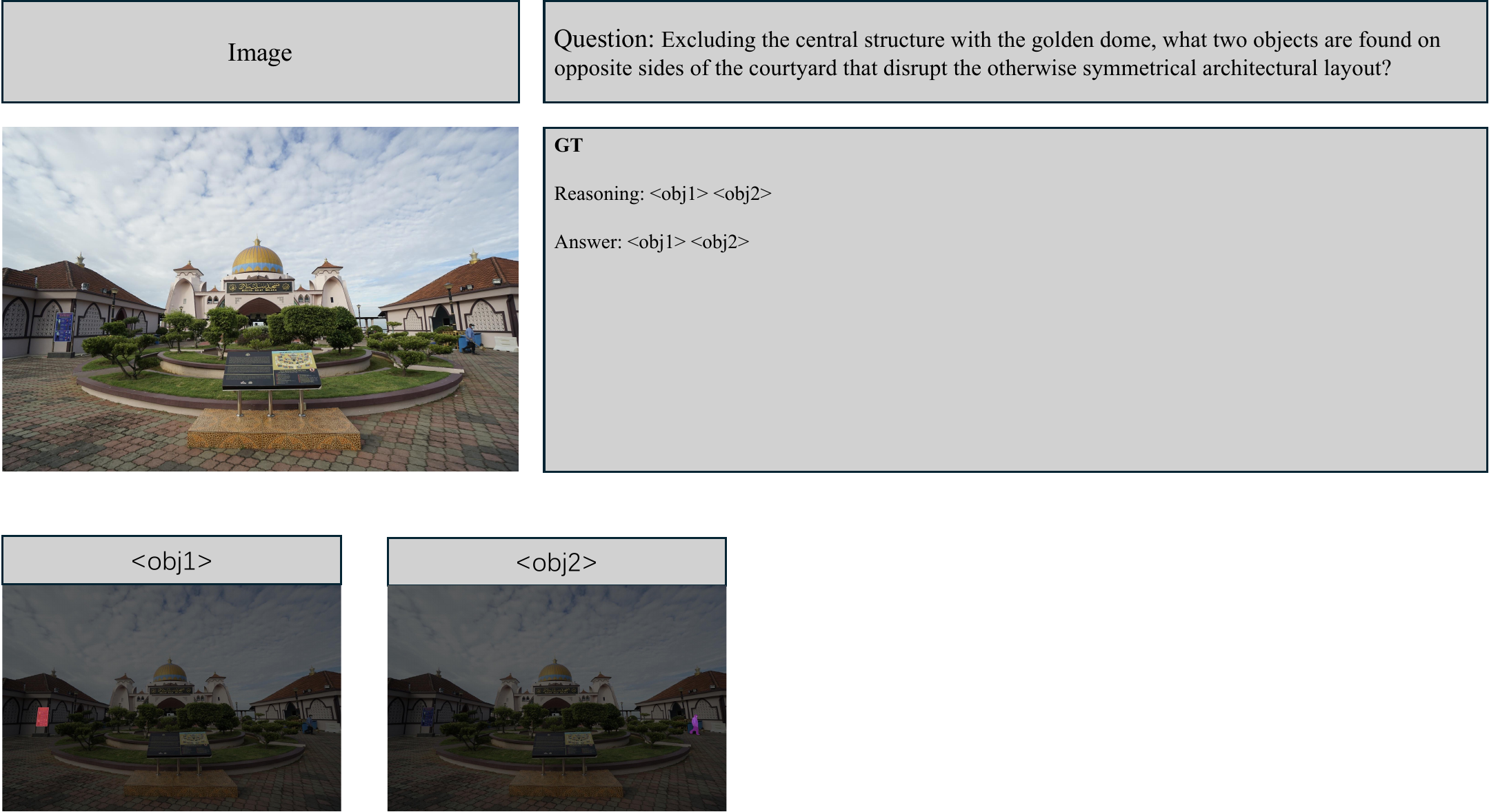}
    \caption{Visualization Results of ground truth.}
    \label{fig:vis_gt2}
    \end{figure*}

As shown in~\Cref{fig:vis_sa1,fig:vis_gp1,fig:vis_q1,fig:vis_gt1}, our R-Sa2VA-Qwen3-4B-RL model produces a significantly more complete and interpretable reasoning trace compared to the Gemini 2.5 Pro and Qwen3-VL-4B baselines. For the same query, R-Sa2VA sequentially identifies the performer, the pavement region, the soccer ball, and finally the hat. Crucially, each step is grounded by a dedicated segmentation mask (\texttt{<obj1>}–\texttt{<obj4>}), clearly linking textual reasoning to the corresponding visual traces. In contrast, while Qwen3-VL-4B-Instruct+SAM2 generates a plausible step-by-step explanation, it only segments the final answer, leaving intermediate cues (person, ball, and pavement) ungrounded. Similarly, Gemini 2.5 Pro+SAM2 primarily outputs a detection-style description of the hat, localizing the final object but lacking an explicit multi-object reasoning path. These visualizations demonstrate that while existing MLLM+SAM2 pipelines can often locate the target, our model exposes a coherent, object-level visual reasoning process. A second example is presented in~\Cref{fig:vis_sa2,fig:vis_gp2,fig:vis_q2,fig:vis_gt2}. This scenario requires both robust spatial awareness and fine-grained visual feature perception; notably, only our R-Sa2VA model successfully derives the correct final answer. Interestingly, the Qwen3 baseline (\Cref{fig:vis_q2}) correctly infers from language that the answer category is a ``signboard,'' but the predicted box is misaligned and does not correspond to the ground-truth signboard, revealing a mismatch between its textual inference and spatial grounding.

\section{Discussion of VRT-Bench}\label{sec:diss}

Our VRT-Bench is mainly focusing on the localizing the target object during the reasoning process. Here we discuss and compare with other related works.

\noindent\textbf{MMR}~\cite{jang2025mmr}
focuses on multi-target and multi-granularity reasoning segmentation: a single instruction may refer to multiple objects and parts, and the dataset provides 194K complex, often implicit queries over object- and part-level masks. In contrast, VRT is not primarily about multi-target or part reasoning, but about making the entire reasoning path observable and evaluable. Each VRT sample contains a question, a step-by-step textual explanation and, crucially, a segmentation mask for every mentioned object along the reasoning path, plus the final answer mask. VRT-Bench further groups samples into comparison, function, location, and visual-detail categories and introduces Logic Quality (LQ) and Visual Quality (VQ) to measure how well models recover the sequence of grounded objects, rather than only the final target.

\noindent\textbf{MIRA}~\cite{zhou2025visualizing}
proposes a benchmark for visual chain-of-thought where intermediate visualizations (sketches, structural diagrams, path drawings) are necessary for solving complex reasoning problems, and evaluates models under text-only, text-plus-image, and full visual-CoT settings. VRT shares the spirit of “visualized reasoning”, but the representation and domain are quite different. MIRA asks models to generate imagined visual images that guide reasoning; VRT instead uses pixel-accurate segmentation masks on real images as the visual traces, aligning each textual step with a set of object masks. In short, MIRA studies how imagined sketches help reasoning; VRT studies how object-level segmentation traces can be used to evaluate and train grounded reasoning.

\noindent\textbf{Seg-Zero}~\cite{liu2025seg}
is a method that decouples reasoning and segmentation: a reasoning model generates explicit chain-of-thought and positional prompts, and a segmentation model uses these prompts to produce masks. The system is trained purely with GRPO-style reinforcement learning, using format and accuracy rewards, and achieves strong zero-shot performance on ReasonSeg~\cite{lai2024lisa} without any explicit supervision of reasoning traces. Our framework R-Sa2VA is orthogonal and complementary. We also leverage RL after the SFT, but VRT provides ground-truth multi-step traces. In practice, Seg-Zero demonstrates that RL can elicit emergent reasoning without explicit trace labels (but only with language reasoning), whereas VRT offers a benchmark and dataset for quantitatively evaluating such methods at the reasoning-trace level (LQ/VQ), not only by final segmentation IoU.

\noindent\textbf{PRIST}~\cite{cai2025pixel}
is a benchmark where multi-turn conversations drive pixel-level segmentation; the MIRAS framework tracks evolving user intent and produces pixel-grounded explanations together with target masks. Compared to PRIST, VRT currently focuses on single-turn image–question pairs rather than conversational interaction, but provides more structured supervision on the reasoning path: every intermediate object mentioned in the explanation carries a segmentation mask and is evaluated by LQ/VQ. Pixel-level RS emphasizes dialog and intent tracking; VRT emphasizes fine-grained, object-level decomposition of a single reasoning problem. The two directions are complementary: PRIST could in principle adopt VRT-style metrics for multi-turn traces, while VRT could be extended to conversational settings.

\noindent\textbf{RSVP}~\cite{lu2025rsvp}
is a method that uses a two-stage pipeline: multimodal chain-of-thought visual prompts first generate interpretable region proposals, which are then refined by a vision–language segmentation module; this improves final results. RSVP thus focuses on designing a reasoning-aware segmentation model, but the training and evaluation still revolve around traditional segmentation metrics and final masks. VRT, by contrast, proposes a task, benchmark, and dataset where the intermediate visual reasoning trace is the central object of evaluation: models are asked to output a structured sequence of (text step, mask set) pairs, and we separately score logical coverage and spatial accuracy. We view RSVP-style architectures as natural candidates to be re-trained and re-evaluated on VRT-80k and VRT-Bench, using our trace metrics as additional evaluation.

\section{\textbf{Future Work}}\label{sec:future}

\begin{figure*}
        \centering
    \includegraphics[width=.7\linewidth]{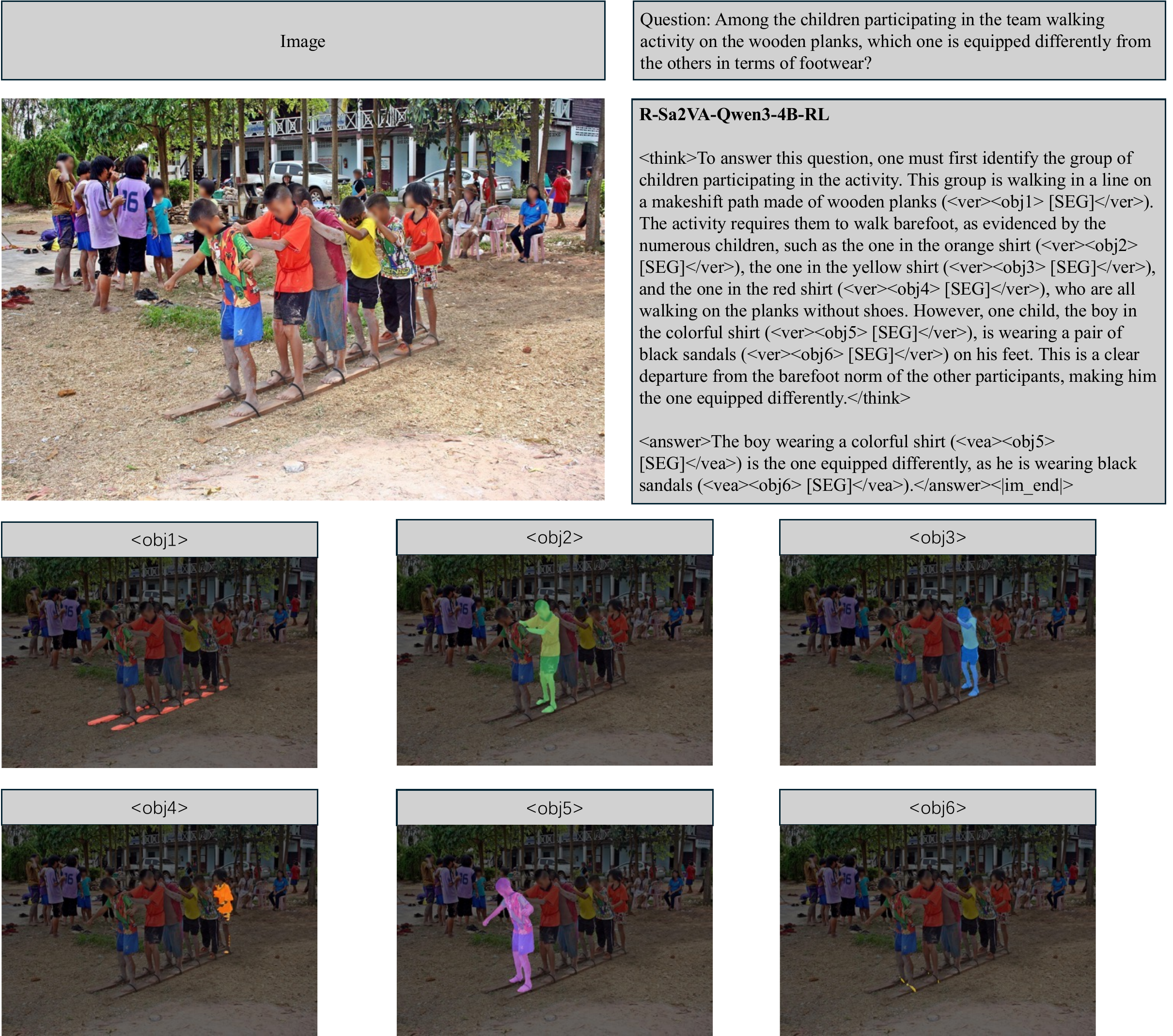}
    \caption{Visualization of failure case of R-Sa2VA-Qwen3-VL-4B-RL.}
    \label{fig:vis_sa2va_f1}
    \end{figure*}

\begin{figure*}
        \centering
    \includegraphics[width=.7\linewidth]{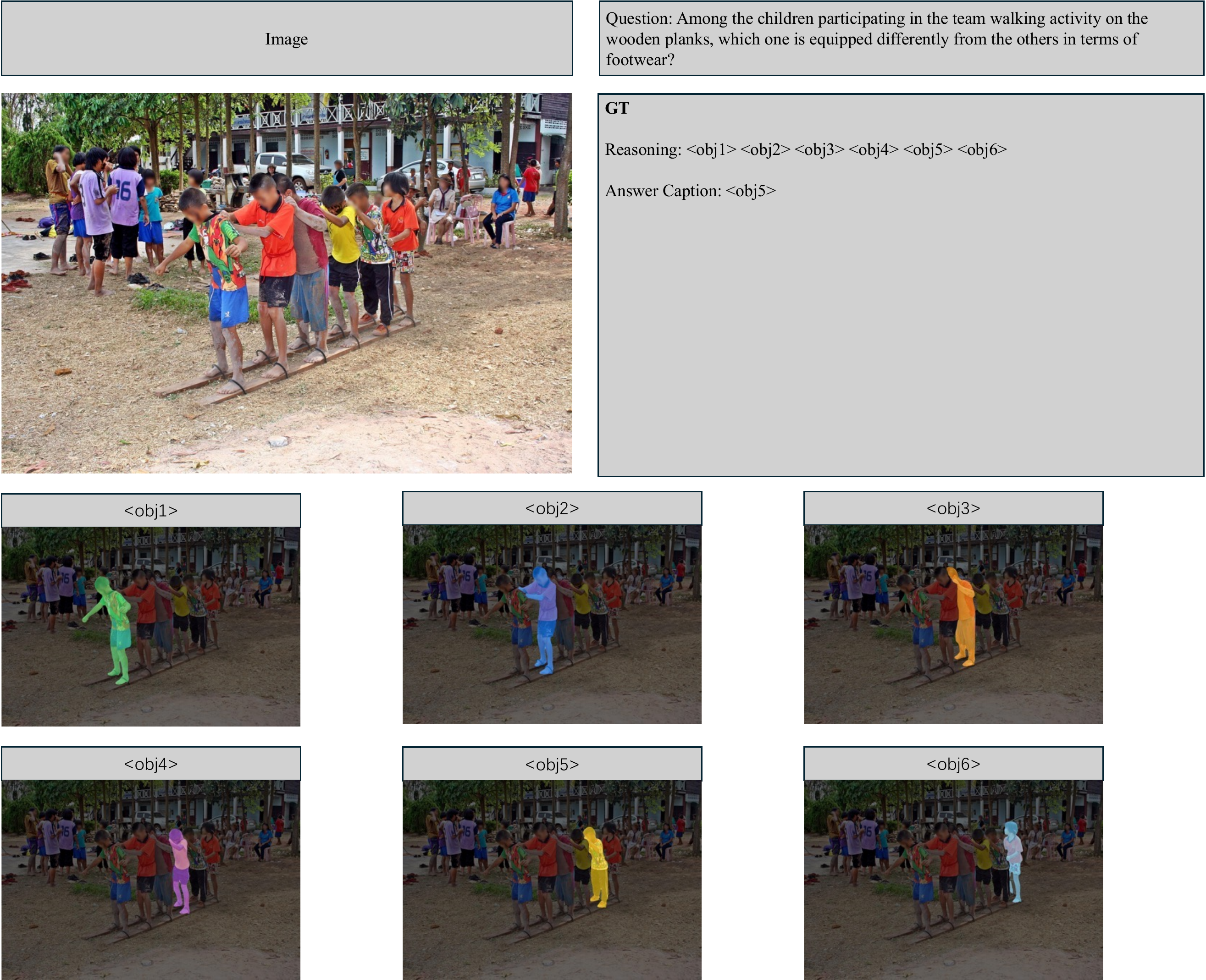}
    \caption{Visualization of ground truth.}
    \label{fig:vis_gt_f1}
    \end{figure*}

\noindent\textbf{Failure Cases.} Despite R-Sa2VA's strong visual reasoning tracing ability and performance on VRT-Bench, there are still several failure cases need to be discussion. As illustrated in \Cref{fig:vis_sa2va_f1} and \Cref{fig:vis_gt_f1}, the image depicts a group of children participating in a team-building activity, standing on two long wooden boards and attempting to walk together. The question requires identifying which child is equipped differently in terms of footwear. In this specific example, R-Sa2VA's reasoning process involves segmenting the wooden planks(\texttt{<obj1>}) and referencing multiple children(\texttt{<obj2>}-\texttt{<obj5>}) based on their shirt colors. While these objects offer contextual support, they are not the core objects required to trace the final answer exhaustively based on the footwear comparison. Finally, R-Sa2VA fails to segment the most important child wearing shoes, instead mistaking the black laces(\texttt{<obj6>}) of the wooden planks for sandals.

\begin{figure*}
        \centering
    \includegraphics[width=.7\linewidth]{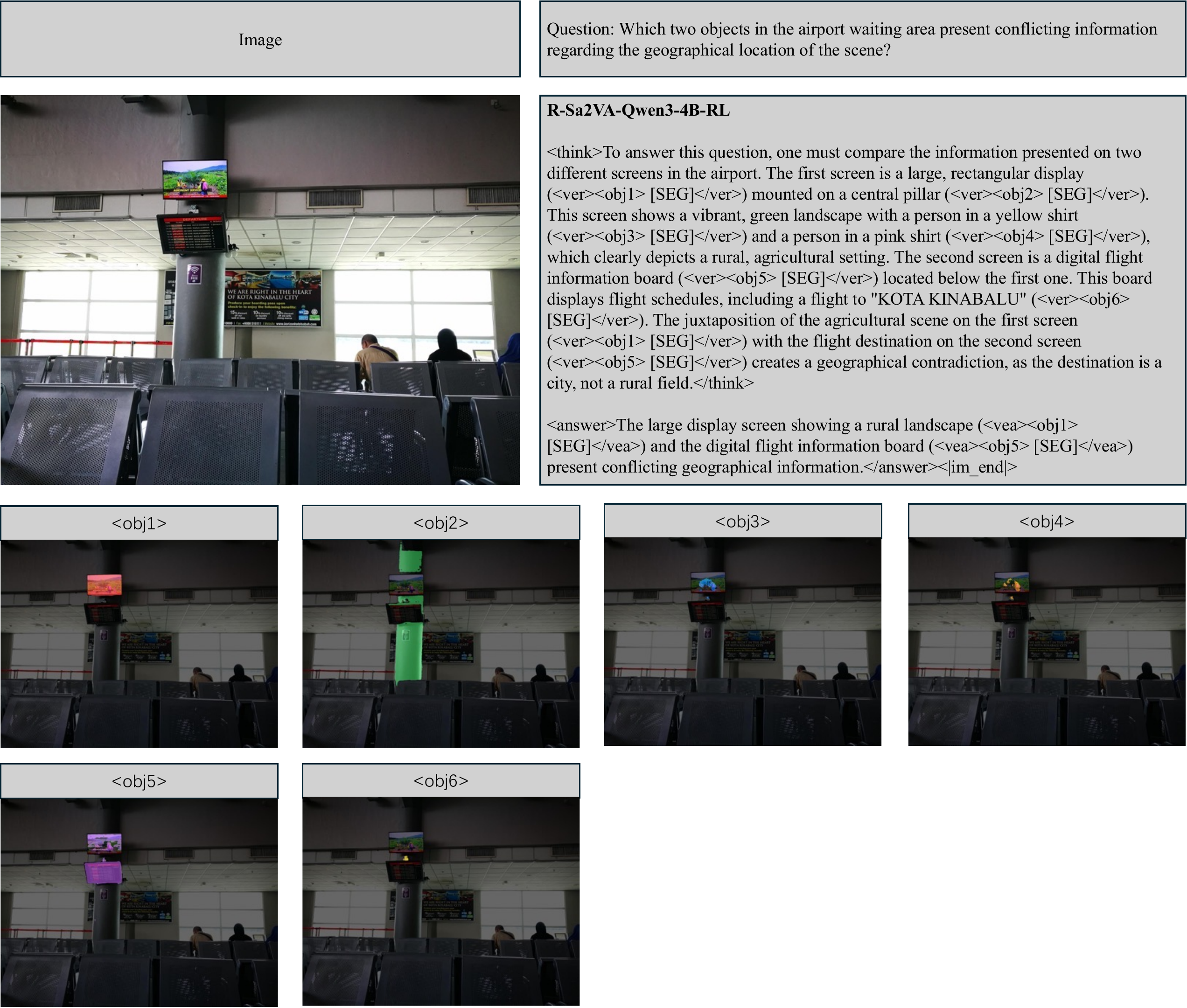}
    \caption{Visualization of failure case of R-Sa2VA-Qwen3-VL-4B-RL.}
    \label{fig:vis_sa2va_f2}
    \end{figure*}

\begin{figure*}
        \centering
    \includegraphics[width=.7\linewidth]{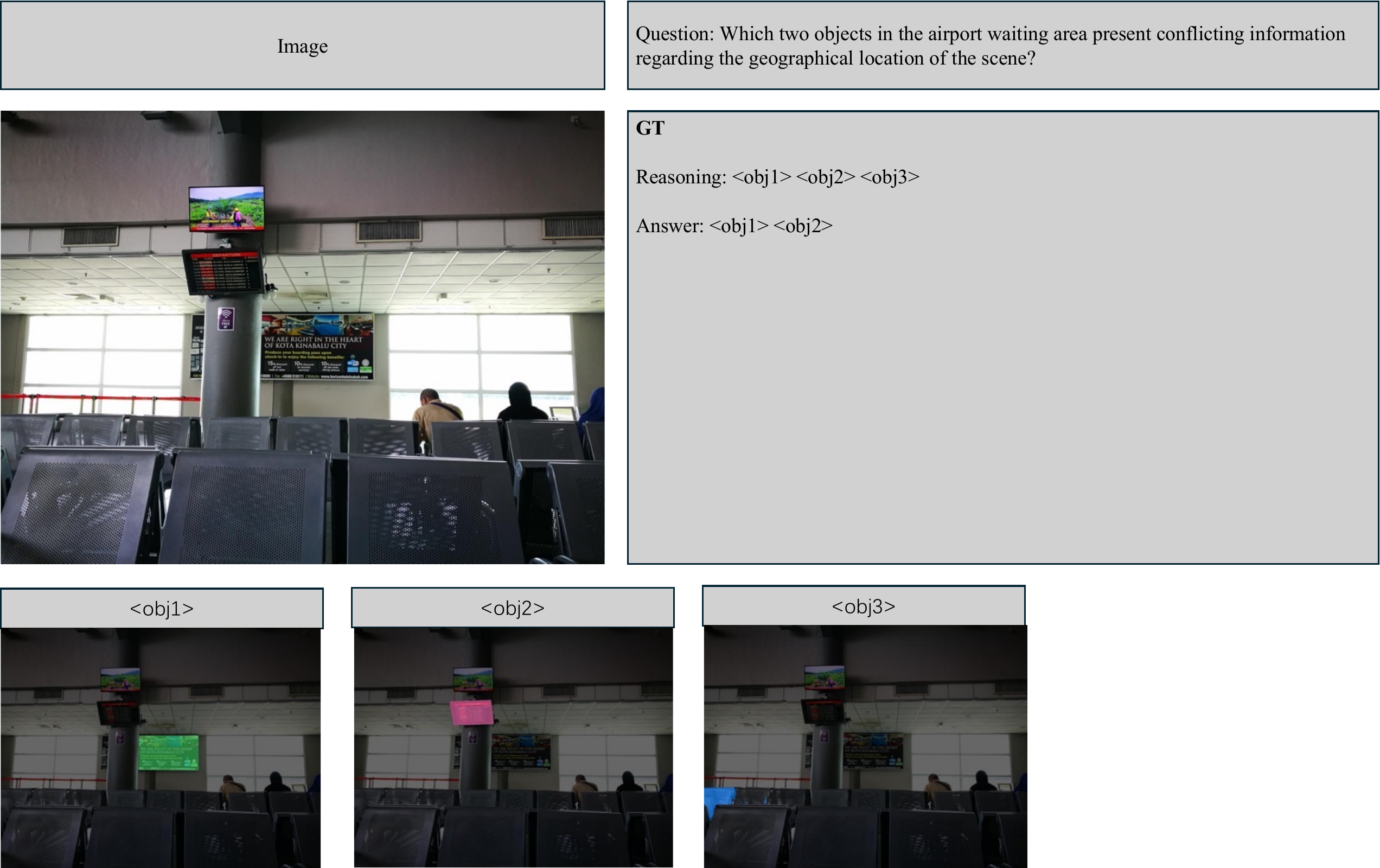}
    \caption{Visualization of ground truth.}
    \label{fig:vis_gt_f2}
    \end{figure*}

\begin{figure*}
    \centering
    \includegraphics[width=1.\linewidth]{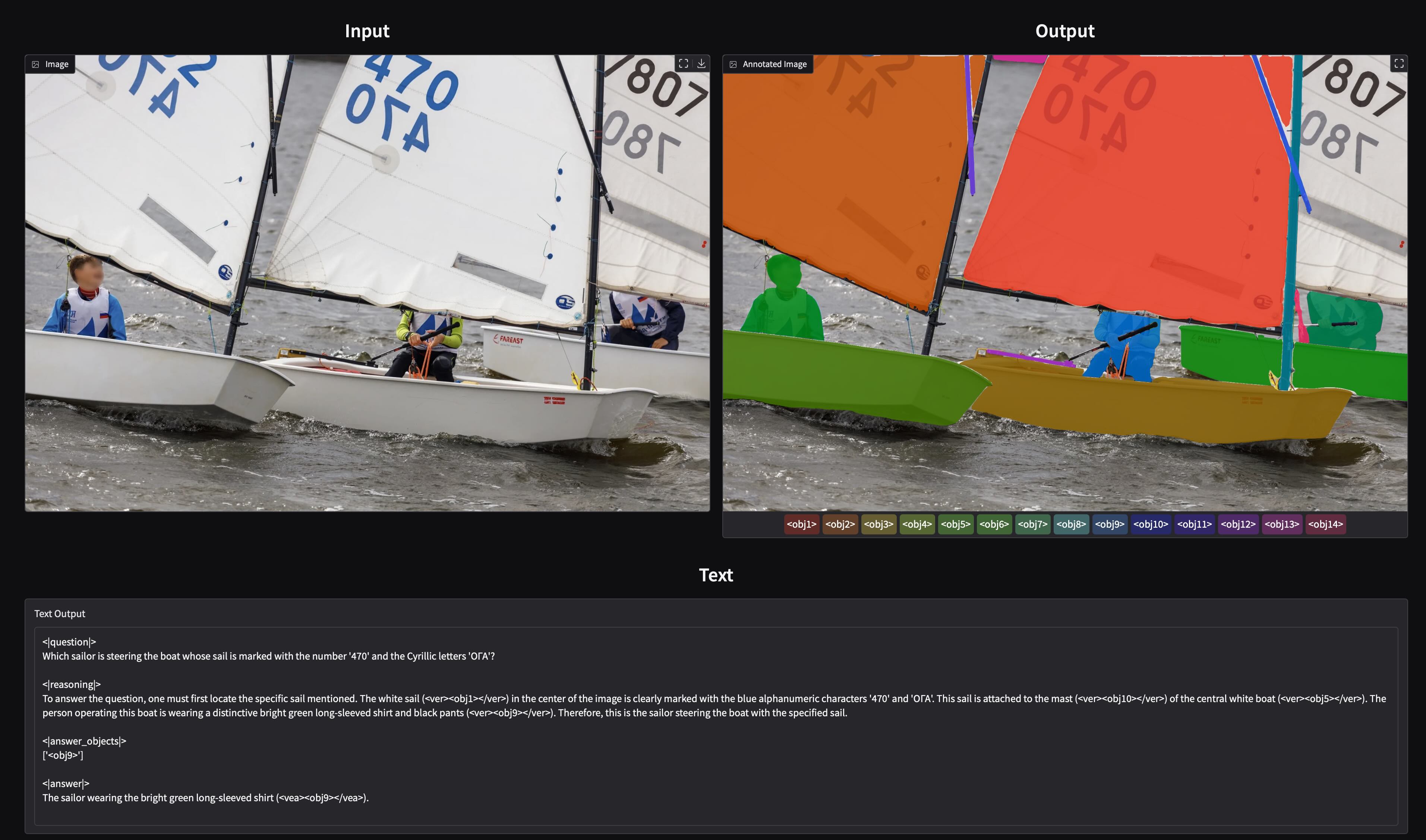}
    \caption{Our Labeling Tool. }
    \label{fig:labeling}
\end{figure*}

\begin{figure*}
    \centering
    \includegraphics[width=1.\linewidth]{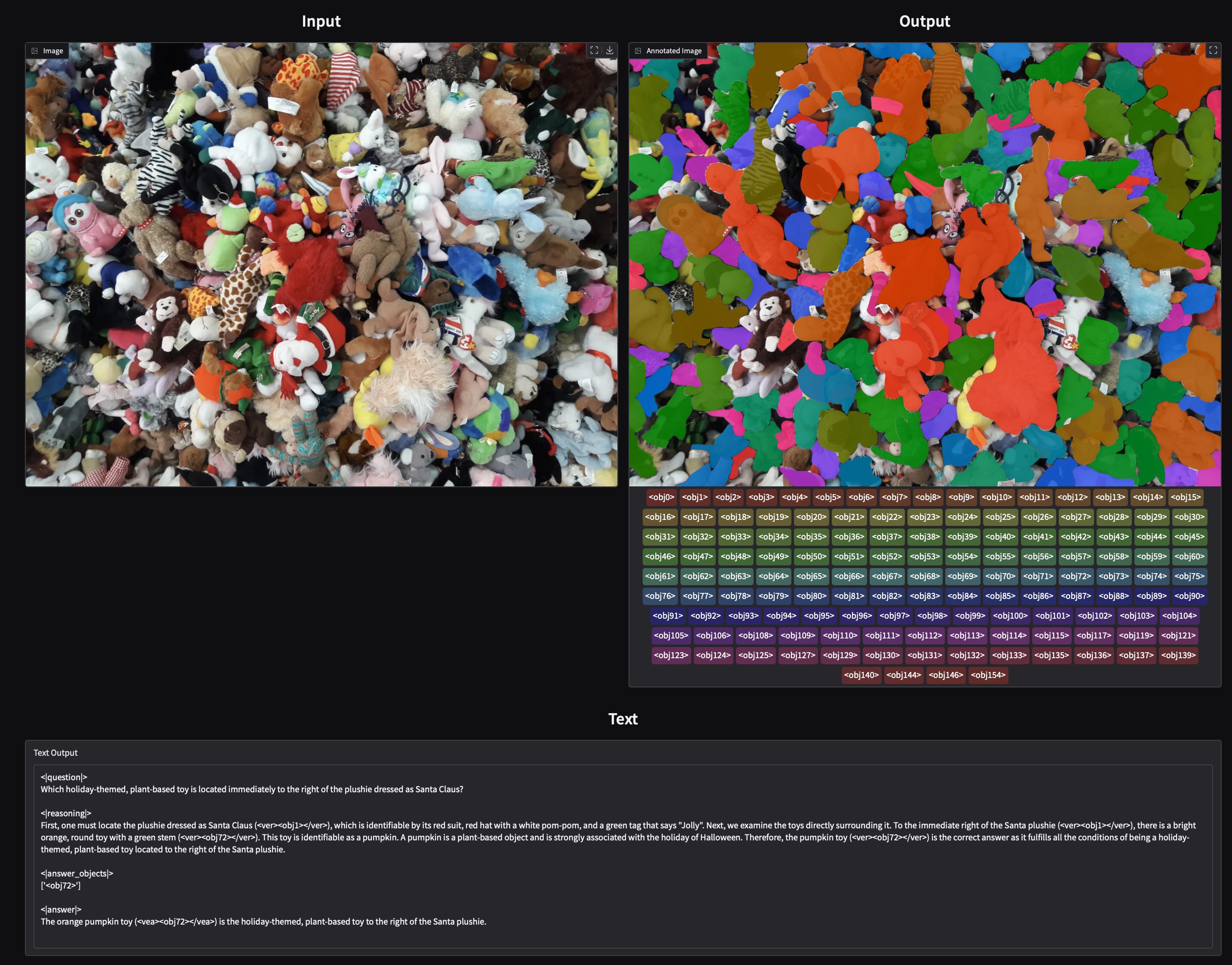}
    \caption{Filtered Example 1.}
    \label{fig:fail1}
\end{figure*}

\begin{figure*}
    \centering
    \includegraphics[width=1.\linewidth]{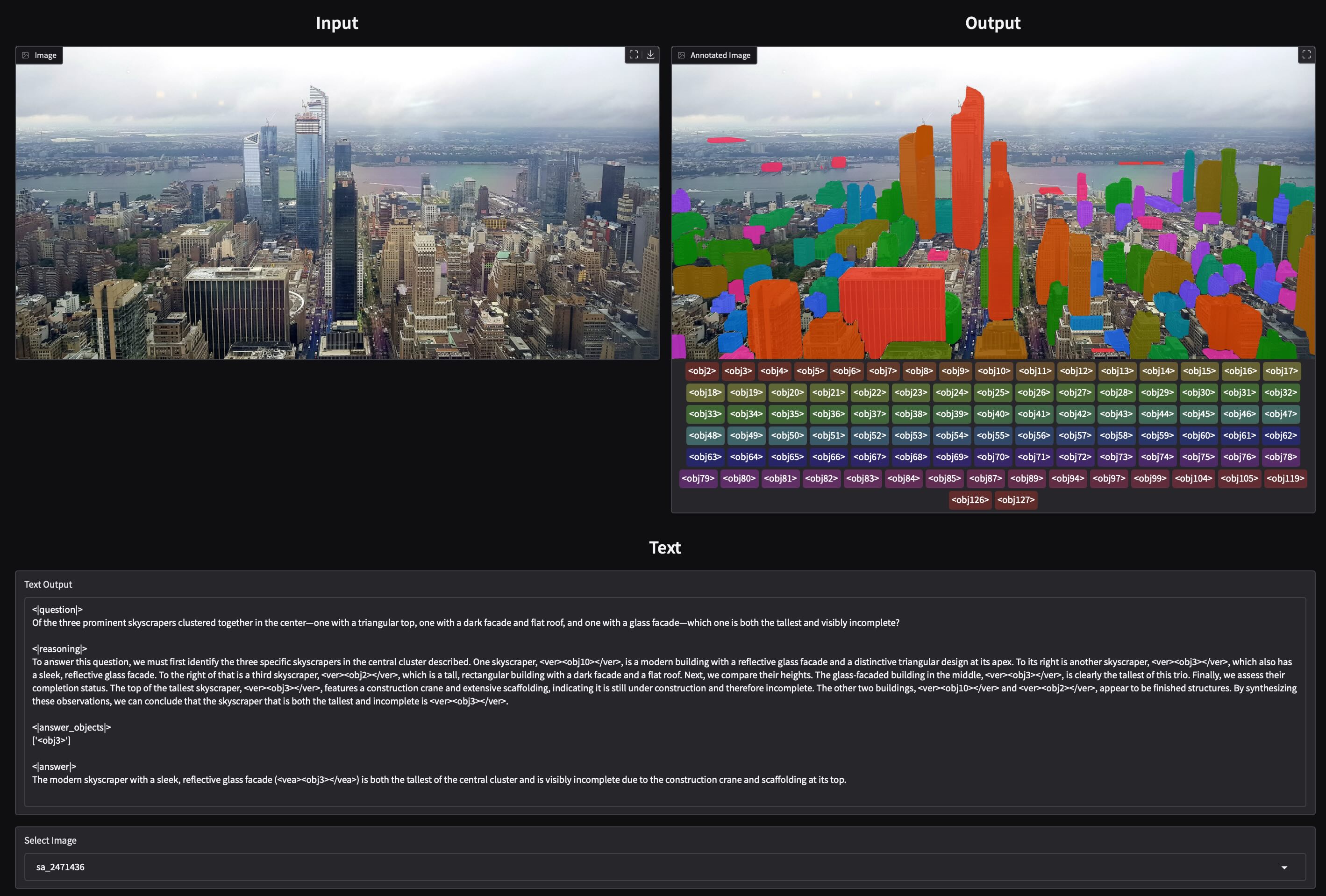}
    \caption{Filtered Example 2.}
    \label{fig:fail2}
\end{figure*}

Similarly, as shown in \Cref{fig:vis_sa2va_f2} and \Cref{fig:vis_gt_f2}, the image presents an airport waiting area. The question requires identifying two objects that present conflicting information regarding the geographical location. The Ground Truth highlights the contradiction between the digital screen(\texttt{<obj1>}) where 'KOTA KINABALU' is listed as a destination for multiple flights and a wall poster explicitly stating 'WE ARE RIGHT IN THE HEART OF Kota Kinabalu City' which implies that the viewer's current location is Kota Kinabalu. However, R-Sa2VA fails to locate the critical text-based poster in the background and misses the definitive semantic conflict about the scene's location provided by the poster.

It is observed that when the scene contains a dense collection of objects or text-rich background elements, R-Sa2VA's visual reasoning trace tends to focus on prominent features while missing subtle but critical details. This failure highlights that enabling the model to generate a large number of masks to comprehensively cover the scene including small objects and background text is a crucial aspect of improving visual reasoning tracing capability.

\noindent\textbf{Future directions of VRT-Bench.}
While VRT-80k and VRT-Bench provide a first step toward grounded visual reasoning, there are several promising directions for future works. First, it is promising to further expand VRT-80k in both scale and diversity, incorporating richer scenes, longer reasoning chains, and additional data sources (e.g., videos) to better cover real-world visual reasoning scenarios. Second, our current analysis suggests that RL mainly improves the logical structure of the reasoning while offering limited gains in visual quality. Exploring more advanced training strategies, such as unified segmentation and reasoning model, segmentation-aware reward designs, or consistency regularization across reasoning steps, may substantially improve the segmentation quality along the entire reasoning trajectory. Third, future work can build a more comprehensive benchmark by adding more fine-grained reasoning types, expanding to multi-image and temporal settings, and including cross-dataset evaluations, enabling a more systematic study of generalization and failure modes for future visual reasoning tracer models.

% {
%     \small
%     \bibliographystyle{ieeenat_fullname}
%     \bibliography{main}
% }

\newpage
% Visualization Benchmark

% Visualization 1

% Visualization 2

% Visualization failure case


\input{prompt/prompt}

%% file: tabs/tab_abl4_filter.tex
\begin{table*}[t]
\centering
\caption{\small{\textbf{Effect of Data filtering.} We report Logic Quality (\textbf{R-LQ}), Visual Quality (\textbf{R-VQ}), and Answer mIoU (\textbf{A}.)}}
\resizebox{1.\textwidth}{!}{
% Updated the column definition to 'ccc' for each metric group
\begin{tabular}{l|ccc|ccc|ccc|ccc|ccc}
\toprule[0.2em]
% --- Column Headers ---
% \multicolumn now spans 3 columns for each metric
\multirow{2}{*}{Model} & \multicolumn{3}{c|}{\#comp} & \multicolumn{3}{c|}{\#func} & \multicolumn{3}{c|}{\#loc} & \multicolumn{3}{c|}{\#visf} & \multicolumn{3}{c}{Overall} \\
% \cmidrules updated to span 3 columns each (e.g., 2-4, 5-7)
\cmidrule(lr){2-4} \cmidrule(lr){5-7} \cmidrule(lr){8-10} \cmidrule(lr){11-13} \cmidrule(lr){14-16}
% Second header row updated with the new sub-metrics
 & R-LQ & R-VQ & A & R-LQ & R-VQ & A & R-LQ & R-VQ & A & R-LQ & R-VQ & A & R-LQ & R-VQ & A \\
\midrule[0.1em]
\rowcolor{gray!20} R-Sa2VA-Qwen3VL-4B-RL & 67.2 & 86.6 & 63.8 & 69.4 & 87.6 & 62.6 & 65.3 & 85.8 & 57.1 & 65.4 & 86.9 & 63.4 & 67.0 & 86.7 & 62.1\\
- Filter & 66.0 & 87.1 & 63.8 & 71.1 & 87.2 & 63.8 & 65.5 & 85.9 & 56.7 & 65.6 & 87.0 & 62.6 & 67.5 & 86.8 & 61.8\\
\bottomrule[0.1em]
\end{tabular}
}
\label{tab:abl_filter}
\end{table*}

%% file: figs/fig_eval_results_comprehensive.tex
\begin{figure*}[t]
    \centering
    \includegraphics[width=1.\linewidth]{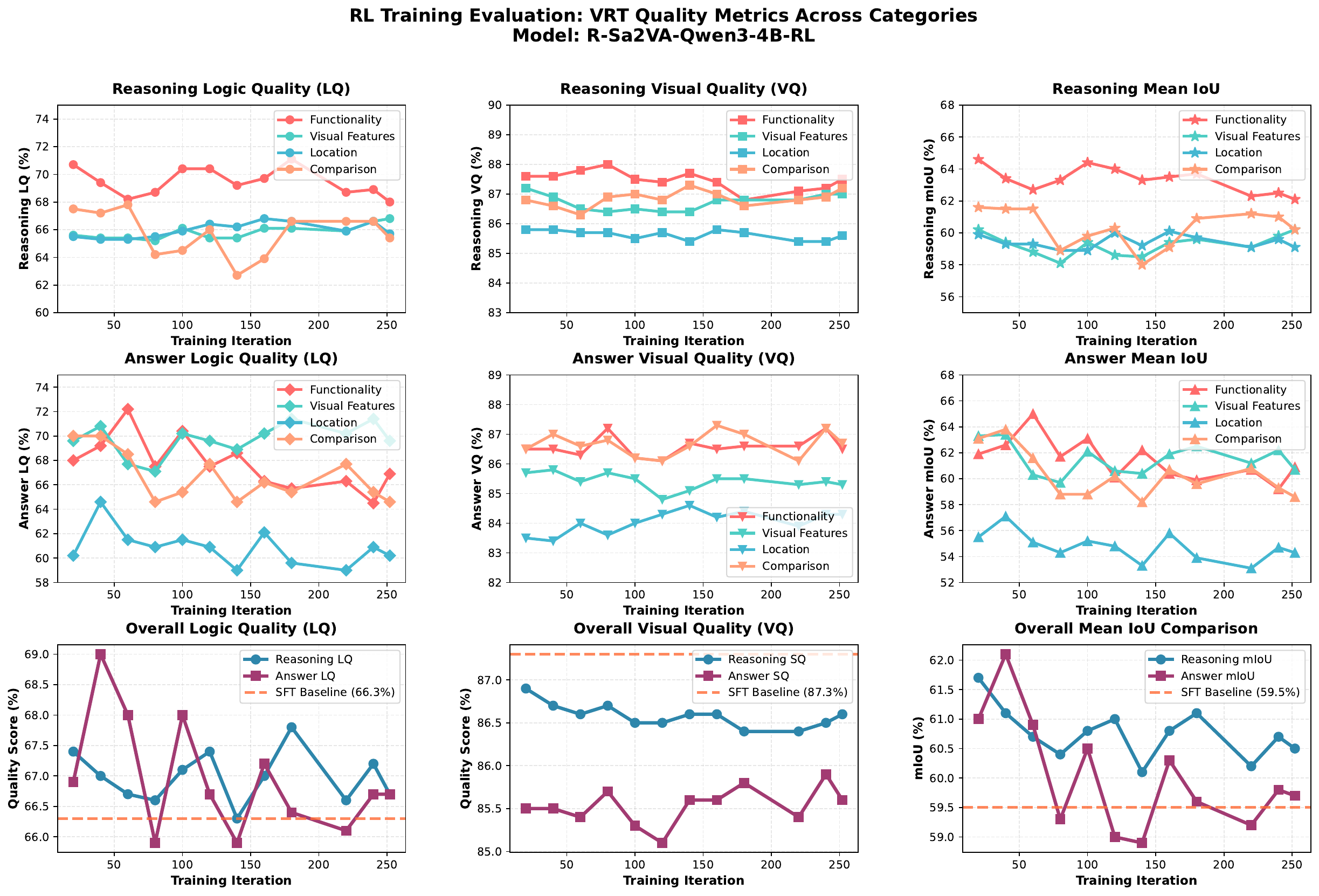}
    \caption{\textbf{RL training dynamics on VRT-Bench.} Evolution of reasoning/answer Logic Quality (LQ), Visual Quality (VQ), and mean IoU (mIoU) during reinforcement learning of R-Sa2VA-Qwen3VL-4B-RL. The first two rows report per-category scores for Functionality, Visual Features, Location, and Comparison, while the bottom row summarizes overall reasoning and answer metrics and compares them to the SFT-only baseline (orange dashed lines).}
    \label{fig:eval_results_comp}
\end{figure*}

%% file: prompt/prompt.tex
\begin{figure*}[t]
    \centering
    \begin{rawprompt}[title=Data Labeling Prompt (Part I)]
**Your Role:** You are an Expert Visual Evidence Reasoning Analyst.

**Your Mission:** Your task is to analyze an image and its corresponding object descriptions to generate a single, high-quality, complex, and **self-contained** question-answer pair. You will provide the question, a step-by-step reasoning process, and the final answer, all formatted into a specific JSON output.

---

### **Guiding Principle: Create a Self-Contained Puzzle**
This is the most important rule. Imagine a human will see **only the image and your final question**. They will **not** see the detailed descriptions you were given. Therefore, your question and reasoning must be answerable from the visual evidence alone. Use the provided descriptions as your "ground truth" to ensure your observations are correct, but do not assume the user has access to them.

---

### **Input You Will Receive**
**Important: IGNORE ANY INFORMATION OF THIS EXAMPLE FOR YOUR ANSWER**

You will be provided with two pieces of information: an image and its detailed object descriptions.

**Description Format Example:** (Image is ignored in example, the provided image is NOT for this example)
```description
<obj0>: The house is a two-story brick structure...
<obj1>: The sky is a clear, vibrant blue...
<obj2>: A large tree with a dense canopy of reddish-brown leaves...
<obj3>: The sidewalk is composed of large, rectangular concrete slabs...
<obj4>: The grass is lush and green...
```

---

### **Core Task: Generating the Question-Answer Pair**

Your process should follow these steps:

**Step 1: Analyze and Verify**
First, carefully compare the `[DESCRIPTIONS]` with the `[IMAGE]`. The descriptions are your factual basis. If they are fundamentally inaccurate or misrepresent the image content, stop the task. Report the mismatch in the `comments` field of your output. If they match, proceed.

**Step 2: Formulate a Complex Question (`question`)**
Craft a single, insightful question in clear English. Your question **must**:
*   **Be Concise:** The question should be a single, direct sentence. **AVOID** long preambles and **AVOID** multi-part questions.
*   **Require Synthesis:** It should necessitate reasoning about the relationships between **multiple objects**.
*   **Go Beyond Identification:** Explore relationships that are spatial, ranking, functional, social, causal, or hypothetical.
*   **Be Grounded:** The question must be directly answerable using *only* one or more of the provided objects (`[DESCRIPTIONS]`) in the image.

**Step 3: Construct the Reasoning (`reasoning`)**
Provide a step-by-step logical explanation detailing how to arrive at the answer.
*   **Chain of Logic:** Your reasoning must clearly walk through the evidence, step-by-step, showing how you synthesized information from different objects.
*   **Explicit Referencing:** Every object mentioned in the reasoning **must** be referenced by its ID (e.g., `<obj0>`, `<obj2>`).
*   **Describe, Don't Quote:** The reasoning should describe the visual properties of the objects, not quote the descriptions you were given. For example, instead of writing "The description of `<obj0>` says it is red," you must write "`<obj0>` is red." Your reasoning should read as a visual analysis of the image itself. Remember, you are creating self-contained puzzle, you cannot see any `[DESCRIPTIONS]` when solving the puzzle.
*   **Visual Evidence Reasoning Tags:** In your reasoning, wrap ALL object references with `<ver><objN></ver>` tags. This includes both answer objects and supporting evidence objects. Label as many relevant objects as possible to show comprehensive visual analysis.
    \end{rawprompt}
    \caption{Data Labeling Prompt (Part I).}\label{fig:prompt1}
\end{figure*}

\begin{figure*}[t]
    \centering
    \begin{rawprompt}[title=Data Labeling Prompt (Part II)]
**Step 4: Provide the Answer**
*   **`answer_objects`:** Create a list containing **only the unique IDs** of the object(s) that are the direct answer to your question.
*   **`answer_caption`:** Write a brief language answer connecting the objects.
*   **Visual Evidence Answer Tags:** In your `answer_caption`, wrap ONLY the answer objects with `<vea><objN></vea>` tags. Do not tag supporting evidence objects that are not part of the direct answer. For example, if the answer is "a person wearing a red shirt," and only the person is the answer object, use: "A person wearing a red shirt (<vea><obj1></vea>)" - do NOT tag the shirt as a separate answer object unless it's specifically part of the answer.
*   **Answer Equivalence:** Either `answer_objects` or `answer_caption` can be directly used as the answer.

**Step 5: Self-Critique and Score (`score`)**
Internally score your generated question-answer pair on a scale of 1 to 10. **Only generate a final output for candidates with a score of 7 or higher.** Use these criteria:
*   **Reasoning Complexity (1-10):** How many objects were synthesized? How non-obvious is the connection between them?
*   **Question Quality (1-10):** Is the question insightful, unambiguous, concise, and thought-provoking?
*   **Answer Accuracy (1-10):** Is the answer precisely identified, correct, and directly supported by the reasoning?

---

### **Tag Usage Summary**
*   **Reasoning**: Use `<ver><objN></ver>` for ALL objects mentioned (comprehensive visual analysis)
*   **Answer**: Use `<vea><objN></vea>` for ONLY the direct answer objects (selective answer marking)
*   **Same Objects**: `<ver><obj1></ver>` and `<vea><obj1></vea>` refer to the same object, just used in different contexts
*   **No Other Tags**: Do NOT use any other tags in your output. Do NOT use <ver> in the `answer_caption` and do NOT use <vea> in the `reasoning`.

---

### **Output Format**

Your final output must be a single JSON object.

**A) If you generate a suitable question (score >= 7):**

```json
{
  "candidates": [
    {
      "id": 1,
      "question": "The question you formulated.",
      "reasoning": "Your step-by-step logical explanation, referencing objects like <ver><obj0></ver> and <ver><obj1></ver>.",
      "answer_objects": ["<objN>", "..."],
      "answer_caption": "Text answer connecting the objects. All objects (\"<vea><objN></vea>\", ..,) should be in the text.",
      "score": 8
    }
  ],
  "comments": "The description matches the image. The question requires complex, multi-object reasoning and has a clear, well-supported answer."
}
```
    \end{rawprompt}
    \caption{Data Labeling Prompt (Part II).}\label{fig:prompt2}
\end{figure*}

\begin{figure*}[t]
    \centering
    \begin{rawprompt}[title=Data Labeling Prompt (Part III)]
**B) If no suitable question can be formed (score < 7) or the description is inaccurate:**

```json
{
  "candidates": [],
  "comments": "A suitable question could not be generated because [your reason here, e.g., the objects lack meaningful relationships, the reasoning is too simple, or the description does not fit the image]."
}
```

---

### **Example**

**Note:** The following example is based on the hypothetical descriptions provided earlier. In a real task, you would be viewing the corresponding image.

```json
{
  "candidates": [
    {
      "id": 1,
      "question": "Which object's state suggests a different season than the others?",
      "reasoning": "To determine the likely season, we synthesize clues from multiple objects. The tree (<ver><obj2></ver>) has reddish-brown leaves, and the sky (<ver><obj1></ver>) is a clear, vibrant blue, both of which are strong indicators of a crisp autumn day. However, the grass (<ver><obj4></ver>) is lush and green with a few scattered yellow flowers, a condition more typical of spring or late summer. Therefore, the grass (<ver><obj4></ver>) presents evidence that contradicts the autumn season suggested by the other objects.",
      "answer_objects": ["<obj4>"], # This is the object that suggests a different season. No tag here.
      "answer_caption": "The grass (<vea><obj4></vea>) suggests a different season than the other objects.",
      "score": 9
    }
  ],
  "comments": "The question requires identifying a primary trend (the season) from multiple objects and then performing a secondary reasoning step to find a contradiction. This involves high-level synthesis and evaluation."
}
```
---

### **Input**

Descriptions: 
```text
{DESCRIPTIONS}
```
Please generate the output based on the descriptions and the provided image.
    \end{rawprompt}
    \caption{Data Labeling Prompt (Part III).}\label{fig:prompt3}
\end{figure*}